%% file: main.tex
\pdfoutput=1

\documentclass[11pt]{article}

\usepackage[final]{acl}

\usepackage{times}
\usepackage{latexsym}

\usepackage[T1]{fontenc}

\usepackage[utf8]{inputenc}

\usepackage{microtype}
\usepackage[compact]{titlesec}

\usepackage{inconsolata}

\usepackage{graphicx}

\usepackage{enumitem}
\usepackage{array}
\usepackage{amsthm}
\usepackage{amssymb}
\usepackage{booktabs} 

\usepackage{listings}
\usepackage{longtable}
\usepackage{booktabs}
\usepackage{soul}
\sethlcolor{yellow}
\usepackage{subcaption}
\usepackage{multirow}
\usepackage{bm}
\usepackage{svg}
\usepackage{calligra}

\usepackage{tikz}

\definecolor{taborange}{rgb}{1.0, 0.498, 0.0549}
\definecolor{tabblue}{rgb}{0.1216, 0.4667, 0.7059}

 \usepackage[most]{tcolorbox}

\newtcolorbox{myquotebox}{
  colback=white!0, 
  colframe=black, 
  rounded corners,
  boxrule=0.5pt, 
  title=Prompt:,
  left=2mm, 
  right=2mm, 
  top=1mm, 
  bottom=1mm 
}

\usepackage{todonotes}
\usepackage{xspace}

\definecolor{lightgrey}{RGB}{158, 158, 158}
\definecolor{goldenrod}{rgb}{0,0,0.8}
\definecolor{deepred}{rgb}{0.6,0,0}
\definecolor{deepgreen}{rgb}{0,0.5,0}
\definecolor{pink}{RGB}{219, 48, 122}
\definecolor{forestgreen}{RGB}{34,139,34}
\definecolor{goldenrod}{RGB}{218,165,32}
\definecolor{sepia}{RGB}{112,66,20}

\usepackage[capitalise]{cleveref}
\crefname{figure}{Fig.}{Figs.}
\crefname{table}{Table}{Tables}
\crefname{appendix}{App.}{App.}
\crefname{section}{§}{§§}
\crefname{equation}{Eq.}{Eqs.}

\definecolor{lightred}{RGB}{254, 138, 138}
\definecolor{lightblue}{RGB}{176, 195, 248}
\definecolor{lightgreen}{RGB}{138, 218, 174}
\definecolor{cbgreen}{HTML}{009E73}
\definecolor{cbred}{HTML}{D55E00}
\newcommand\myparagraph[1]{
\vskip 0.05in 
\noindent{\bf {#1}}}

\definecolor{boxborder}{RGB}{86, 113, 209}  
\definecolor{boxbg}{RGB}{255, 255, 255}    
\definecolor{boxtitle}{RGB}{255, 255, 255} 
\definecolor{boxheader}{RGB}{86, 113, 209}  

\newtheoremstyle{compactdef}
  {\topsep}   
  {\topsep}   
  {\itshape}  
  {}          
  {\bfseries} 
  {.}         
  { }         
  {}          

\theoremstyle{compactdef}

\usepackage{bbm}
\newcommand{\ind}{\mathbbm{1}}  

\title{Can Reasoning Help Large Language Models Capture \\Human Annotator Disagreement?}

\author{
 \textbf{Jingwei Ni\textsuperscript{$E$ $Z$}\thanks{Equal contributions.}}\quad
 \textbf{Yu Fan\textsuperscript{$E$}\footnotemark[1]}\quad
 \textbf{Vilém Zouhar\textsuperscript{$E$}}\quad
 \textbf{Donya Rooein\textsuperscript{$B$ $E$}}\quad
\\
 \textbf{Alexander Hoyle\textsuperscript{$E$}}\quad
 \textbf{Mrinmaya Sachan\textsuperscript{$E$}}\quad
 \textbf{Markus Leippold\textsuperscript{$Z$}}\quad \\
 \textbf{Dirk Hovy\textsuperscript{$B$}}\quad 
 \textbf{Elliott Ash\textsuperscript{$E$}}\quad
\\
 \textsuperscript{$E$}ETH Zürich\quad
 \textsuperscript{$Z$}University of Zürich\quad
 \textsuperscript{$B$}Bocconi University
\\
 \texttt{\{\href{mailto:jingni@ethz.ch}{\color{black} jingni}, \href{mailto:yufan@ethz.ch}{\color{black} yufan}, \href{mailto:ashe@ethz.ch}{\color{black} ashe}\}@ethz.ch}
}
  
\begin{document}
\maketitle
\begin{abstract}



\end{abstract}
 Variation in human annotation (i.e., disagreements) is common in NLP, often reflecting important information like task subjectivity and sample ambiguity. Modeling this variation is important for applications that are sensitive to such information. Although RLVR-style reasoning (Reinforcement Learning with Verifiable Rewards) has improved Large Language Model (LLM) performance on many tasks, it remains unclear whether such reasoning enables LLMs to capture informative variation in human annotation. In this work, we evaluate the influence of different reasoning settings on LLM disagreement modeling. We systematically evaluate each reasoning setting across model sizes, distribution expression methods, and steering methods, resulting in 60 experimental setups across 3 tasks. 
 Surprisingly, our results show that RLVR-style reasoning degrades performance in disagreement modeling, while naive Chain-of-Thought (CoT) reasoning improves the performance of RLHF  LLMs (RL from human feedback). These findings underscore the potential risk of replacing human annotators with reasoning LLMs, especially when disagreements are important.
 \footnote{Code and data at \url{https://github.com/EdisonNi-hku/Disagreement_Prediction}.}

\section{Introduction}
Inter-annotator disagreement is common in NLP annotations \citep{snow-etal-2008-cheap} and often treated as noise to be removed by majority voting \citep{SABOU14.497} or expert aggregation \citep{hovy-etal-2013-learning}. However, these solutions may be misguided, as annotation disagreement can signal a diversity of views and often contains valuable information that enables downstream applications to capture a diversity of human values and interpretations \citep{plank-2022-problem}. Human annotators have access to different information sets and are guided by different value systems \citep{fornaciari-etal-2021-beyond, fuchs2021value}. It is therefore not surprising that different annotators give different answers, in particular for subjective tasks such as hate speech detection \citep[e.g.][]{kennedy2018gab} where disagreement often arises from varying sociodemographic and cultural backgrounds \citep{fleisig-etal-2023-majority}.
Even seemingly ``objective'' labeling tasks, such as part-of-speech (POS) tagging, show disagreement due to ambiguous language 
\citep{plank-etal-2014-linguistically,jiang-marneffe-2022-investigating}. Generally speaking, disagreement is natural, contains valuable information, and should not be ignored or erased, but actively modeled \citep{Uma2021LearningFD,leonardelli-etal-2023-semeval}.

With the rapid growth of LLMs' capability, evaluating LLMs' ability to capture annotation disagreement is becoming increasingly important. On one hand, more ``capable'' LLMs achieve better performance in predicting the majority-voted label, and are thus widely adopted to replace human decision-making in applications such as text classification \citep{Pangakis2023AutomatedAW,Trnberg2024BestPF,gpt4_vs_crowdworker}, chatbot preference annotation \citep{lee2024rlaifvsrlhfscaling}, and LLM-as-a-judge \citep{calderon2025alternativeannotatortestllmasajudge,fan2025lexam}. On the other hand, many of these applications also require understanding the full spectrum of annotator disagreement. However, evaluations typically focus on majority-label prediction, overlooking the modeling of underlying disagreement distributions. As a result, it remains unclear whether the LLMs can reliably automate these applications, by effectively flagging cases with potential annotator disagreement for human oversight.  

Prior work evaluates early LLMs and identifies their limitations in modeling annotation disagreement under specific settings \citep{lee-etal-2023-large}, but have largely overlooked several key factors influencing distribution modeling, such as (1) in-context steering methods (e.g., few-shot learning); and (2) distribution expression methods \citep{meister2024benchmarkingdistributionalalignmentlarge}. More importantly, the role of reasoning---which significantly enhances LLM performance in various tasks \citep{wei2023chainofthoughtpromptingelicitsreasoning,deepseekai2025deepseekr1incentivizingreasoningcapability}---is underexplored in prior work \citep{lee-etal-2023-large,chen-etal-2024-seeing}. Presumably, reasoning can benefit disagreement modeling by enabling LLMs to explore and compare different opinions through CoT. However, reasoning may harm decision making when the problem has hard-to-articulate criteria \citep{nordgren2009devil,liu2024mindstepbystep}. This may be particularly relevant to RLVR LLMs, which are optimized on tasks with single-deterministic answers---contrasting with the reality that many tasks involve multiple valid perspectives.

To address these gaps, we conduct a comprehensive evaluation of LLMs under different reasoning settings: RLHF LLM with and without CoT, as well as RLVR LLM. Given that the impact of reasoning may be further influenced by other factors such as LLM size, distribution expression, and steering method \citep{meister2024benchmarkingdistributionalalignmentlarge}, our evaluation systematically explores the full combinations of (1) 3 reasoning settings; (2) 5 LLM sizes (from 8B to 671B); (3) with or without few-shot steering; and (4) 2 distribution expression methods \citep{tian2023justaskcalibrationstrategies,wei2024measuringshortformfactualitylarge}, resulting in 60 prompting settings. We evaluate all settings on 5 datasets of 3 widely studied tasks, following the metrics in prior work: (1) \textit{variance correlation} (VarCorr, \citealp{davani-etal-2022-dealing}), measuring how well the LLM-predicted variance correlates to human annotation variance; and (2) \textit{distributional alignment} (DistAlign, \citealp{meister-etal-2024-towards}), directly comparing the distributional divergence of LLM and human labels.

Surprisingly, we find that RLVR-style reasoning significantly harms disagreement modeling when human annotation variance is high. Moreover, forcing additional reasoning effort \citep{muennighoff2025s1simpletesttimescaling} does not improve the performance of RLVR LLMs. In contrast, for RLHF LLMs, CoT prompting significantly improves disagreement modeling. Furthermore, RLVR LLMs are better with a \textit{deterministic} goal (e.g., predicting the majority annotation) than with a \textit{probabilistic} goal (e.g., predicting the proportion of human disagreements). Our findings suggest that using RLVR-optimized LLMs in disagreement-matter tasks requires extra caution, as these models may overlook critical human disagreements. In summary, our contributions are:

\begin{enumerate}[left=0mm,itemsep=0pt,topsep=1pt]
\item We systematically evaluate RLVR and RLHF LLMs in disagreement modeling across 3 tasks, 5 LLM sizes, and 12 prompting settings.
\item We quantitatively reveal the limitations of RLVR-style reasoning in modeling disagreement (\cref{sec:reasoning}), and provide qualitative insights to explain these findings (\cref{sec:qualitative}). 
\item Our evaluation further examines the impact offers other relevant factors on disagreement modeling, including distribution expression methods (\cref{sec:verb_vs_sample}), the importance of human annotations (\cref{sec:data_importance}), few-shot steering (\cref{sec:few-shot}), and model scale (\cref{sec:scaling}).
\end{enumerate}

\newcommand{\hlc}[2][yellow]{{%
    \colorlet{foo}{#1}%
    \sethlcolor{foo}\hl{#2}}%
}
\definecolor{task-blue}{HTML}{c1e5f5}
\definecolor{task-yellow}{HTML}{fbe3d6}
\definecolor{task-green}{HTML}{c2f1c8}
\definecolor{task-pink}{HTML}{f2cfee}

\begin{figure*}[t]
    \centering
	\includegraphics[width=\linewidth]{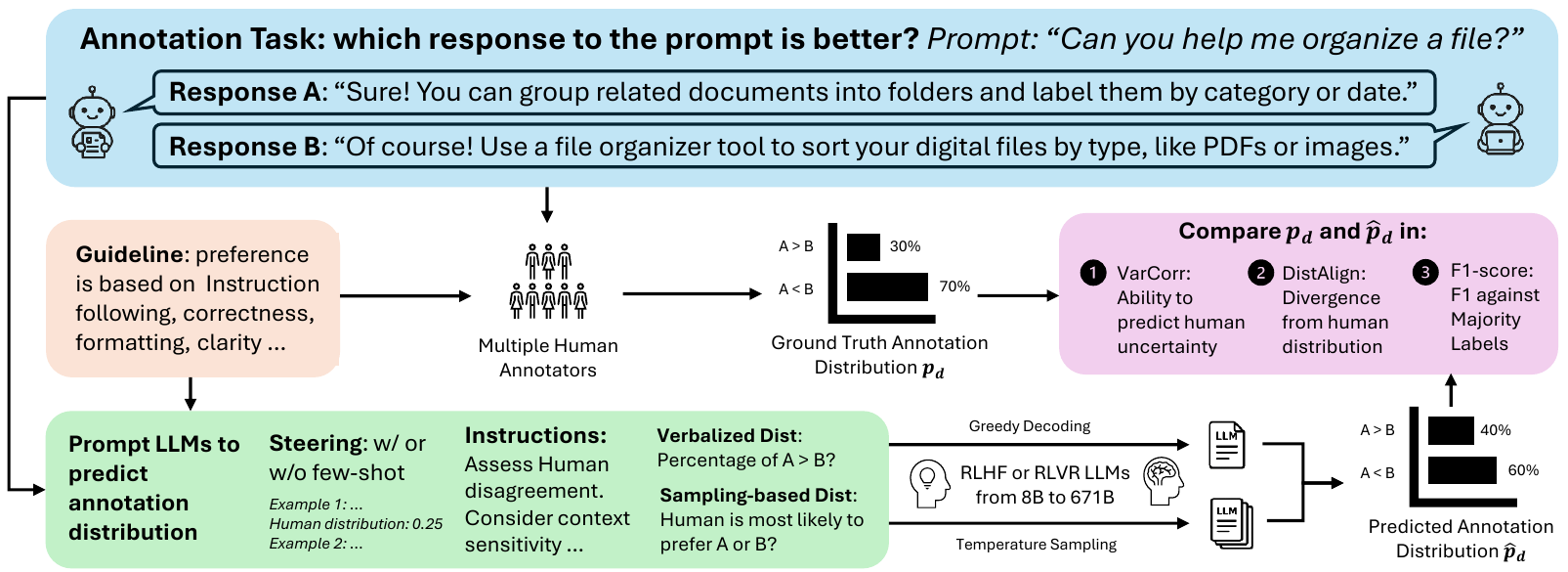}

	\caption{An illustration of our evaluation: We start with a \hlc[task-blue]{task} with \hlc[task-yellow]{guidelines} for both human and \hlc[task-green]{LLM} annotators.
    The LLM predictions of the annotation distributions are then \hlc[task-pink]{compared} with true human label distribution.}
	\label{fig:overview}
\end{figure*}

\section{Background and Related Work} \label{sec:related_work}
\myparagraph{RLHF and RLVR} are two dominant paradigms for LLM alignment. RLHF fine-tunes models using reward models trained on human preference data, optimized via reinforcement learning algorithms like PPO \citep{ouyang2022traininglanguagemodelsfollow}. RLVR instead derives rewards from automatically verifiable properties—such as code execution correctness, passing unit tests, or satisfying mathematical constraints \citep{deepseekai2025deepseekr1incentivizingreasoningcapability}. Intuitively, RLHF prioritizes subjective human preference, while RLVR emphasizes objective problem-solving verification.

\myparagraph{Annotation Disagreement in NLP.} Annotation disagreement has been an important area of study with long history \citep{wiebe-etal-2004-learning,ovesdotter-alm-2011-subjective,basile-etal-2021-need,Uma2021LearningFD,leonardelli-etal-2023-semeval}. 
Various qualitative and quantitative analyses show that the majority of disagreement is caused by other systematic reasons (e.g., ambiguity, context sensitivity etc.) rather than random annotation noise (e.g., carelessness) \citep{plank-etal-2014-linguistically,popovic-2021-agree,jiang-marneffe-2022-investigating,santy-etal-2023-nlpositionality,zhang2024divergingpreferencesannotatorsdisagree}. 

Prior work in modeling disagreement has fruitfully leveraged datasets with annotator metadata (e.g., annotator ID, explanations, and sociodemographic features), enabling annotator modeling and deeper insights into sources of variation \citep{davani-etal-2022-dealing,hu-collier-2024-quantifying,giorgi-etal-2024-modeling,chen-etal-2024-seeing,chochlakis-etal-2025-aggregation,orlikowski2025demographicsfinetuninglargelanguage}. Our evaluation is complementary: we focus on settings where annotator metadata are absent – an increasingly common scenario in large-scale or emergent tasks (e.g., LLMs as judges or annotators, \citealp{cui2024ultrafeedbackboostinglanguagemodels}), and assess how well models can capture disagreement in such constrained but realistic contexts. 

\myparagraph{Distribution Prediction with LLM.} The extensive training corpus of LLMs may enable them to simulate different opinions and predict distribution in real-world \citep{grossmann2023ai,ziems-etal-2024-large}, and numerous previous studies use LLMs to predict the distribution of political opinions \citep{argyle2023out,durmus2024measuringrepresentationsubjectiveglobal,meister2024benchmarkingdistributionalalignmentlarge,karanjai2025synthesizingpublicopinionsllms}. The closest prior work to ours is \citet{lee-etal-2023-large}, which reveals LLMs' limited performance on disagreement modeling for Natural Language Inference (NLI). Specifically, they prompt LLMs to predict NLI labels and probe the annotation distribution with the log probabilities of LLM outputs and Monti-Carlo Sampling outcome. However, their evaluation does not fully address several key aspects: (1) Distribution expression methods based on token-level probability and sampling are shown to be ineffective by \citet{meister2024benchmarkingdistributionalalignmentlarge} (and our results in \cref{sec:verb_vs_sample}); (2) \citet{lee-etal-2023-large} prompt LLMs answer without explicitly instructing LLMs to consider potential disagreement or controversy. Such task-instruction mismatch may hinder LLMs' ability in disagreement modeling; and (3) their study does not investigate the role of reasoning, which can be crucial for LLMs to explore various aspects of disagreements. It also does not consider other factors such as few-shot steering. To address these gaps, we investigate the impact of reasoning with detailed instruction for disagreement modeling, while also examine the influence of distribution expression methods, few-shot steering, and LLM size.

\section{Problem Formalization}

In this section, we formalize the problem of predicting human annotation disagreement and visualize it in \cref{fig:overview}. Let $d \in D$ be a datapoint from a dataset $D$, for which we have a set of $n$ annotations $\mathbf{A_{d}} = \{a_{d,i} | a_{d,i} \in \{0,1\}, i \in \{1, 2, ..., n\}\}$ from different human annotators, indicating if $d$ is a positive (1) or negative (0) sample.\footnote{For simplicity, we study the binary classification problem. Multi-label classification problem with $m$ labels is equivalent to $m$ binary classification problems.}
We assume that the $n$ annotators are representative of the annotator population, so human annotation on $d$ follows a Bernoulli distribution $H_d$ parameterized by:
\begin{equation}
    p_{d} = \frac{|\{a_{d,i} = 1 | a_{d,i} \in \mathbf{A_{d}}\}|}{n}
\end{equation}
where $p_{d}$ denotes the probability that a human annotator labels $d$ positive. The variance of human annotation is $\sigma_d^2 = p_{d}(1 - p_{d})$. 

Given human disagreement as the gold label, a machine learning algorithm is tasked with simulating and predicting it. Specifically, through techniques such as fine-tuning, prompting, or sampling, a model can predict a Bernoulli distribution $\hat{H}_d$ regarding how likely a human will annotate $d$ positive, parameterized by $\hat{p}_{d}$. 
Then, the variance of the machine-predicted annotation is $\hat{\sigma}_d^2 = \hat{p}_{d}(1 - \hat{p}_{d})$.

To evaluate the model's annotation distribution against humans',  we employ two dimensions of evaluation from prior work:

\myparagraph{Variance Correlation.} In automatic annotation, it is crucial for LLMs to identify samples that are likely to elicit disagreements between human annotators. To evaluate this ability, we adopt the variance correlation metric from \citet{davani-etal-2022-dealing}, which quantifies to what extent higher model uncertainty indicates higher human uncertainty. The formula is:
\begin{equation}
    \text{VarCorr} = \text{Corr}\left(\langle\sigma_d^2\rangle_{d \in D}, \langle\hat{\sigma}_d^2\rangle_{d \in D} \right)
\end{equation}
where Corr denotes the Pearson's Correlation \citep{pearson1895note}. 

\myparagraph{Distributional Alignment.} Although VarCorr captures the alignment of uncertainty, it fails to capture the exact gap between the annotation distributions. For example, if $\langle p_{d}\rangle_{d \in D}=\langle 0.4, 0.5\rangle$ and $\langle \hat{p}_{d}\rangle_{d \in D}=\langle 0.1, 0.2\rangle$, the model achieves perfect VarCorr but underestimates the human disagreement. Similarly, $\langle p_{d}, \hat{p}_{d}\rangle =\langle 0.2, 0.8\rangle $ shares the same variance, but has contradictory distribution. Therefore, we adopt Distributional Alignment from \citet{meister2024benchmarkingdistributionalalignmentlarge}, formalized by:
\begin{equation}
    \text{DistAlign} = \frac{1}{|D|} \sum_{d \in D} \lVert p_{d} - \hat{p}_{d} \rVert_1
\end{equation}
which measures the exact difference between two distributions. 
Importantly, DistAlign cannot fully substitute VarCorr in evaluating uncertainty. For example, given the gold labels of samples $\langle p_{1}, p_{2}\rangle=\langle 0.33, 0.4\rangle $, model prediction (A) $\langle \hat{p}_{1}, \hat{p}_{2}\rangle =\langle 0.4, 0.33\rangle$ is better than (B) $\langle \hat{p}_{1}, \hat{p}_{2}\rangle=\langle 0.15, 0.4\rangle$ in DistAlign.
However, (B) has better VarCorr than (A) and correlates better with human uncertainty.

Therefore, both VarCorr and DistAlign are important dimensions to evaluate the prediction of disagreement. 

\myparagraph{F1 on Majority Label.} LLMs (especially with RLVR) are optimized to predict the majority labels. Therefore, we adopt F1-score to study the difference between disagreement modeling and majority label prediction. Specifically, we compute $\text{F1}(\langle\ind\{p_d > 0.5\}\rangle_{d \in D}, \langle\ind\{\hat{p}_d > 0.5\}\rangle_{d \in D})$ where $\ind$ is the indicator function. We drop data points with $p_d$ or $\hat{p}_d$ equal to $0.5$ to avoid biased tie-break.

\section{Datasets}
Hate speech detection \citep{warner-hirschberg-2012-detecting,waseem-2016-racist} and emotion classification \citep{hirschberg2003experiments,mihalcea2006corpus} are two broadly studied tasks in annotation disagreement. We follow \citet{davani-etal-2022-dealing} and include Gab Hate Corpus (hereafter GHC; \citealp{kennedy2018gab}) and GoEmotions \citep{demszky-etal-2020-goemotions} for our evaluation. GoEmotion is a multi-label classification dataset. We divide it into three binary classification problems---annotating whether a post contains (1) positive / negative / ambiguous emotions, or not (0). GoEmotion Subtasks hereafter referred to as Pos, Neg, and Amb.  
Furthermore, we include HelpSteer2 (hereafter HS2; \citealp{wang2025helpsteer2preferencecomplementingratingspreferences}), which consists of multiple annotators' preferences for the helpfulness of chatbot responses. Therefore, our evaluation includes five datasets: hate speech detection, chatbot preference classification, and classifications of positive, negative, and ambiguous emotions.

We further derive two subsets of interest from the dataset of each task: (1) \texttt{Random} subset: a randomly sampled subset with 1k data points; and (2) \texttt{HighVar} subset: a subset of 200\footnote{Size of \texttt{HighVar} is determined by the limited number of data points with at least two disagreements. The size of \texttt{Random} is determined for budget control.} data points where at least two annotators disagree with the majority label, and where the overall proportion of the minority label ($1 - p_d$) falls between $\frac{1}{3}$ and  $\frac{1}{2}$ to ensure high annotation variance. \texttt{Random} keeps the original data distribution, containing a lot of samples where human achieves agreement and certain samples where human disagrees. It is useful for evaluating VarCorr---how a model is helpful in predicting human annotation variance. \texttt{HighVar} contains samples with potential systematic disagreement (e.g., two annotators disagree with the other three). Therefore, it is useful in evaluating DistAlign---when there exist separate opinions, can a model detect that and predict an aligned distribution? Dataset preparation details can be found in \cref{appendix:data_preparation}.

Notably, we do not evaluate F1 and VarCorr on \texttt{HighVar}, as predicting majority labels or annotation variance is ill-defined when human annotators already exhibit high annotation variance.

\myparagraph{Low Annotation Noise.} Annotators' carelessness may lead to divergent labels, instead of systematic disagreements. To reduce such noise, we keep data points with more than 3 annotations for evaluated subsets. For the \texttt{HighVar} subsets, there should be at least two annotators disagree with the majority, where the disagreement is less likely due to annotation noise \citep{sandri-etal-2023-dont}. Results in \cref{sec:data_importance} also suggest that our evaluation datasets contain predictable systematic disagreement.

\section{Methodology}

We first motivate our evaluation design in \cref{sec:meta_analyses}. Then we describe the implementation and prompt details in \cref{sec:implementation_details}.

\subsection{Evaluation Motivations and Design} \label{sec:meta_analyses}

\myparagraph{Worth Exploring Factors in Distribution Prediction.} We start by identifying factors that may affect disagreement modeling, but was not addressed in prior work \citep{lee-etal-2023-large,chen-etal-2024-seeing}. (1) \textbf{Distribution Expression Methods}: we can probe prediction distribution from LLMs by either directly asking for a verbalized probability, or by sampling multiple LLM responses and using the answer frequency as the probability (see math formulas for verbalized and sampling-based distribution in \cref{appendix:distribution_formula}). Some previous work find the former more effective \citep{tian2023justaskcalibrationstrategies,meister2024benchmarkingdistributionalalignmentlarge} while others have contradictory observations \citep{wei2024measuringshortformfactualitylarge}. (2) \textbf{In-Context Steering}: In-context steering methods provide LLMs with specific target group information to enhance distribution prediction. \citet{meister2024benchmarkingdistributionalalignmentlarge} find few-shot steering enhances opinion simulation, but its role in disagreement modeling remains underexplored. 

\myparagraph{Evaluate Combinations of Different Factors.} Factors like distribution expression, steering, and LLM size can impact both reasoning and disagreement modeling. To estimate the causal effect of reasoning on disagreement modeling, it is necessary to evaluate all combinations of these factors (i.e., potential confounders) with different reasoning settings. Otherwise, for example, an observed effect of reasoning under a sampling-based distribution method \citep[e.g.,][]{lee-etal-2023-large} may not generalize to verbalized distribution methods. See \cref{app:causal} for detailed causality theories that motivate our design.

\subsection{Implementation Details} \label{sec:implementation_details}

\myparagraph{Prompt-Based Methods.} We evaluate three reasoning settings (RLHF LLMs w/ or w/o CoT, or using RLVR LLMs instead) across the combinations of promising settings discussed in the previous section---namely, (1) with or without few-shot steering; (2) verbalized or sampling-based distribution. Hence, there are $3\times2\times2=12$ settings to be evaluated in total.

To make RLHF and RLVR LLMs comparable, we use DeepSeek-R1 series LLMs \citep{deepseekai2025deepseekr1incentivizingreasoningcapability} (e.g., DeepSeek-R1-Distill-Llama-70B) and corresponding RLHF LLMs sharing the same base LLM (e.g., Llama-3.3-70B-Instruct). To investigate the effect of scaling in LLM size, we experiment LLMs of 8B, 14B, 32B, 70B, and 671B parameters\footnote{We exclude 7B LLMs because their base LLM, Qwen2.5-7B-Math, is specialized for mathematical tasks and therefore unsuitable for the current task.}. 

The prompt structure is illustrated in \cref{fig:overview}. For few-shot illustration, we carefully balance the 5 examples---2 of human-agreed positives and negatives correspondingly, and 1 human-disagreed---to avoid introducing spurious bias \citep{turpin2023languagemodelsdontsay} to distribution prediction. For verbalized probability, we follow \citet{meister2024benchmarkingdistributionalalignmentlarge} to directly ask for the proportion of human annotators that may annotate the sample positive. For sampling-based distributions, we ask for the most likely human label and sampling 10 times with a temperature of 0.7 for conventional LLMs, and 0.6 for reasoning LLMs, following the official recommendation. 

Furthermore, all prompts present LLMs with the same annotation guidelines as in the original dataset papers, which are likely the guidelines presented to human annotators. This may increase LLMs' chance to capture human disagreement caused by the context or natural ambiguity of annotation guidelines. We also explicitly prompt LLMs to assess potential disagreement and consider context sensitivity (e.g., cultural, social, linguistic ambiguity) that may influence the interpretation. Full prompts and inference hyperparameter / budget are detailed in \cref{appendix:prompts} and \cref{appendix:hyperparameter} respectively.

\myparagraph{Fine-tuning Methods.} Fine-tuning encoder-only LMs for disagreement modeling is a straightforward way to use human labels \citep{davani-etal-2022-dealing,fleisig-etal-2023-majority}. Therefore, we fine-tune ModernBERT-large \citep{warner2024smarterbetterfasterlonger} and DeBERTa-V3-large \citep{he2023debertav3improvingdebertausing} to regress onto the positive annotation probability of human $p_{d}$. The loss function is:
\begin{equation}
    \mathcal{L}_{\text{MSE}} = \frac{1}{|D_\mathrm{train}|} \sum_{d\in D_\mathrm{train}} \left( \hat{p}_{d} - p_{d} \right)^2
\end{equation}
where $\hat{p}_{d} = \text{LM}(d)$ is the prediction of the encoder-only LM; and $D_\mathrm{train}$ denotes a randomly sampled training set. Fine-tuning baselines require thousands of data points and repeated human labels to capture the target distribution. This is not applicable for most automatic annotation tasks with limited human labels without majority voting aggregation. Fine-tuning details are in \cref{appendix:ft_detail}.

\section{Results} \label{sec:result}
\begin{table}[t]
\centering
\small
\input{figures/stat_significance}

\caption{\label{tab:stat_sig} Win rates (in \%) of the left settings with Wilcoxon signed-rank tests.
We evaluate on the \texttt{Random} and \texttt{HighVar} subsets.
The intensity of \hlc[Green3!50]{green} and \hlc[Red1!50]{red} indicates how strongly the left setting wins over or loses to the right one.
Statistically significant wins or losses are marked with $^{**}$ ($p<0.01$) and $^{*}$ ($p<0.05$).  }

\end{table}

\begin{table*}[h]
\centering
\resizebox{\textwidth}{!}{
\input{figures/tab_performance_metrics_corrected.tex}
}

\caption{Performance on \texttt{Random} (randomly sampled) subsets of all datasets, aggregating 8B--671B results by Average or Best. Color intensity reflects relative performance within each column. RLVR LLMs shows no significant advantage over RLHF LLMs. 
}
\label{tab:rnd_performance}

\end{table*}

\begin{table}[h]
\centering
\resizebox{\columnwidth}{!}{
\input{figures/tab_disagree_subset_perfomance.tex}
}

\caption{DistAlign Performance on \texttt{HighVar} (high annotation variance) subset of all datasets. RLVR LLMs constantly underperforms RLHF LLMs on both Avg and Best. 
}
\label{tab:dis_performance}
\end{table}

This section presents the evaluation results and takeaways. We start from comparing distribution expression methods---verbalized vs. sampling-based distribution. Then, we investigate the role of reasoning settings and other factors. Due to the large number of experiments, we present aggregated results to convey core messages and present the full model-level performance in \cref{appendix:full_performance}.

\subsection{Verbalizing or Sampling?} \label{sec:verb_vs_sample}
We compare verbalized and sampling-based distributions across 120 controlled experimental settings, varying only the distribution expression method. These settings span 4 LLM sizes (8B, 14B, 32B, and 70B\footnote{We exclude the 671B model due to the high cost of sampling-based prediction.}), 3 reasoning paradigms (RLVR, RLHF with and without CoT), 5 datasets, and 2 steering strategies (few-shot or no steering).

The winning rates of the verbalized distribution in different metrics are shown in the first row of \cref{tab:stat_sig}, combined with the results of the Wilcoxon test \citep{Wilcoxon1992} to show statistical significance. We observe that the verbalized method significantly outperforms in predicting annotation distribution (VarCorr and DistAlign). However, the sampling-based method is better in predicting the majority label (F1). This indicates that predicting the majority label and disagreement are different tasks that require separate evaluations.

\textbf{\textit{Takeaway:}} we recommend evaluating LLM disagreement modeling with verbalized distribution, instead of sampling-based approach in prior work \citep{lee-etal-2023-large}. LLM annotators relying on sampling-based self-consistency to improve majority label prediction may need extra caution, as the sampling-based approach may overlook disagreements \citep[e.g.][]{pangakis2023automatedannotationgenerativeai, ni-etal-2024-afacta, zhou2025selfconsistencyinternalrewardmodels,wang2025creamconsistencyregularizedselfrewarding}. 

Given the significantly better performance of verbalized distribution, we focus the analyses in the following sections on results obtained with this method. Sampling-based methods yield better majority label prediction, which lies outside the scope of disagreement modeling. We therefore analyze those results separately in \cref{appendix:f1_analyses}.

\subsection{Reasoning for Disagreement Modeling} \label{sec:reasoning}
\begin{table*}[ht]
\centering
\resizebox{\textwidth}{!}{
\input{figures/tab_scaling}
}

\caption{Correlation of performance and log-number of LLM parameters ($log(8)$ to $log(671)$). \colorbox{Green3!60}{Green} and \colorbox{Red1!60}{red} intensity reflects the degree of positive / negative scaling.}
\label{tab:scaling}
\end{table*}

We compare reasoning methods---(1) RLHF LLMs without reasoning; (2) RLHF LLMs with CoT reasoning; and (3) lengthy reasoning with RLVR LLMs---across 50 controlled settings, varying only the reasoning methods. Controlled settings span 5 LLM sizes (8B, 14B, 32B, 70B, 671B), 5 datasets, and 2 steering strategies (few-shot or no steering).

Results on \texttt{Random} and \texttt{HighVar} are presented in \cref{tab:rnd_performance} and \cref{tab:dis_performance} respectively. We aggregate the results of 5 LLM sizes by the average and best scores to enable straightforward comparisons between reasoning methods. Rows 2 and 3 of \cref{tab:stat_sig} present the comparisons of (1) RLVR vs. RLHF (w/ or w/o CoT); and (2) RLHF w/ vs. w/o CoT across 50 controlled settings. 

When comparing RLVR LLMs with their RLHF counterparts, we observe that (1) on \texttt{HighVar} where humans strongly disagree with each other, RLVR LLMs achieve significantly worse performance in both aggregated scores in \cref{tab:dis_performance} and setting-level comparisons summarized in \cref{tab:stat_sig}. (2) On \texttt{Random}, results are more mixed but RLVR model does not significantly outperform their RLHF counterparts, as \cref{tab:stat_sig} row 2 shows. However, the \cref{tab:stat_sig} row 3 shows that CoT reasoning in RLHF LLMs improves the performance on both \texttt{Random} and \texttt{HighVar}, compared to without CoT.

To better understand the effect of long reasoning with RLVR LLMs, we force these models to think longer by replacing the end of thinking token ``</think>'' with ``Wait'', which effectively boosts performance for math reasoning \citep{muennighoff2025s1simpletesttimescaling}. We force longer reasoning twice, and compare to the results to natural ending. The controlled comparisons span 40 settings---4 LLM sizes\footnote{We exclude the 671B DeepSeek-R1 since this model is accessed through API, which does not allow forcing longer reasoning}, 2 steering methods, and 5 datasets. The row 4 and 5 of \cref{tab:stat_sig} show the results, where forcing longer reasoning rarely leads to statistically significant improvements. 

Moreover, RLVR underperforms RLHF on majority label prediction (F1) with verbalized distribution as shown by \cref{tab:stat_sig}. However, when applying sampling-based method, RLVR significantly outperforms RLHF on F1 (win rate \colorbox{Green3!15}{62.5\%$^{**}$}). This may be because, in sampling, LLMs are prompted to predict the most likely human label (i.e., majority label), while considering disagreement. This \textit{deterministic} goal is more suitable for RLVR LLMs than the \textit{probabilistic} goal of predicting the proportion of disagreement. However, the sampling-based method still leads to worse distributional prediction as discussed in \cref{sec:verb_vs_sample}.

\textbf{\textit{Takeaway:}} CoT reasoning with RLHF LLMs may benefit the prediction of disagreement. However, people should be more cautious about lengthy reasoning with RLVR LLMs, which can significantly harm the performance in probabilistic disagreement modeling.

\subsection{Human Labels are Important} \label{sec:data_importance}
To study whether it is necessary to gather repeated human labels for disagreement modeling, we compare small LMs – ModernBERT and DeBERTa-V3 – fine-tuned on large-scale human annotations, to the best LLM results. From \cref{tab:rnd_performance} and \cref{tab:dis_performance}, we observe that fine-tuned small encoder-only LMs outperforms LLMs on GHC \texttt{Random}, HS2 \texttt{HighVar}, and all GoEmotions subsets, indicating the value of real human annotations in predicting disagreement. However, LLM-based methods are also promising, achieving better performance on HS2 \texttt{Random} and GHC \texttt{HighVar} without human annotations. 

\textbf{\textit{Takeaway:}} incorporating human labels is highly beneficial for accurate disagreement modeling, while LLM-based methods also demonstrate strong potential due to their cost efficiency and solid performance on certain tasks.

\subsection{Few-Shot Steering} \label{sec:few-shot}
\citet{meister2024benchmarkingdistributionalalignmentlarge} show that LLMs exhibit strong few-shot steerability in distribution prediction. Therefore, we investigate whether few-shot illustrations can steer LLMs for better disagreement modeling. Few-shot is compared to zero-shot prompting across 75 controlled settings---spanning 5 LLM sizes (8B to 671B), 3 reasoning settings, and 5 datasets. Comparisons are summarized in the sixth row of \cref{tab:stat_sig}. Few-shot steering decreases the performance on 4 metrics, with statistically significant drop in 3 of them. 

Observing \cref{tab:rnd_performance} and \cref{tab:dis_performance}, we notice that few-shot steering seems to help certain tasks (e.g., GHC \texttt{Random}) but harm others (e.g., HS2). Therefore, we separately evaluate the effect of few-shot steering on each dataset (see the lower half of \cref{tab:stat_sig} before the last row). The results show that few-shot steering significantly harms disagreement modeling on HS2 and GE-Pos, but improves performance on GHC \texttt{Random} and GE-Neg \texttt{HighVar}. 

\textbf{\textit{Takeaway:}} few-shot steering can be helpful, but its effectiveness varies across tasks and datasets.

We also perform similar per-dataset analyses in earlier sections (e.g., comparing reasoning settings), which mostly yield consistent trends with the aggregated results. We thus only include the aggregated results in \cref{tab:stat_sig} and briefly discuss the per-dataset results in \cref{appendix:perdata}.

\subsection{Scaling Effect of LLM Size} \label{sec:scaling}

Our coverage of LLMs from 8B to 671B allows exploring the scaling effect of LLM size in disagreement modeling. Specifically, we compute the correlation between performance improvement and the increase of log-number of parameters. \cref{tab:scaling} reports the Pearson's coefficients spanning 30 settings---5 datasets, 2 steering methods, and 3 reasoning settings. The comparison across 30 settings are summarized in the last row of \cref{tab:stat_sig}. Scaling LLM size can improve disagreement modeling with statistical significance. However, the improvement is less significant on \texttt{HighVar} while more significant for majority label prediction (F1). \cref{tab:scaling} also shows that different datasets seem to have different scaling effect. Conducting Wilcoxon Test for each dataset, we find that there is statistical significant negative scaling on the disagreement modeling of Neg \texttt{Random}. Other trends are consistent with the results observed across all datasets. 

\textbf{\textit{Takeaway:}} Scaling LLM size may more effectively boost majority label prediction than disagreement modeling. Negative scaling occurs especially in cases of strong disagreement (\texttt{HighVar} subsets) or on specific datasets (e.g., Neg \texttt{Random}).

\subsection{Impact of LLM Size and Steering Method on Reasoning} \label{sec:reasoning_subset}
Will reasoning's effect on disagreement modeling change with different LLM sizes or steering methods? To investigate this, we compare reasoning settings within subsets of conditions where either the steering method or the LLM size is held fixed. Specifically, we evaluate reasoning effects in: (1) all settings with few-shot steering, (2) all settings without few-shot steering, and (3) all settings using specific LLM sizes (e.g., all settings with 8B LLM). Across these subsets, there are no statistically significant observations that contradict those in \cref{sec:reasoning}. Thus, the effect of reasoning remains consistent regardless of the steering method or LLM size.

\subsection{Qualitative Analysis} \label{sec:qualitative}
To understand why RLVR LLMs perform worse than their RLHF counterparts, we conduct a qualitative analysis on GHC and GoEmotions. Specifically, we sample 20 data points from the \texttt{HighVar} subset, and other 20 from \texttt{Random} with low disagreement, focusing on cases where DeepSeek-R1 and V3 have divergent predictions. We find that \textbf{RLVR and RLHF LLMs have different focus of instruction following} although they are prompted exactly the same---In 85\% of cases, RLVR LLMs focus on the annotation guideline, assuming humans would objectively follow the guideline in the same way; while RLHF LLMs focus on considering people with diversified background. One potential reason is that RLVR LLMs are optimized on objective math and coding tasks, thus focusing more on the objective / less controversial parts of prompts. More details and examples in \cref{app:qualitative}.

\section{Conclusion and Discussion}
We evaluate the impact of reasoning on LLM disagreement modeling, with systematic controls of distribution expression, steering, and LLM size. Results show that it requires extra caution to apply RLVR-style reasoning to tasks where annotator disagreements are prevalent and important.

RLHF LLMs exhibit greater potential than RLVR LLMs in predicting disagreements (\cref{sec:reasoning}). This may be because RLVR optimization on verifiable and deterministic answers harms the ability to capture multiple debatable answers. In contrast, reasoning (CoT) with RLHF LLMs improves disagreement modeling, suggesting that the reduced performance of RLVR is not necessarily due to reasoning itself. This may also be related to recent observations that RLVR models can hallucinate more than RLHF models in some tasks \citep{nytimes2025aimorepowerful}. 

Interestingly, \citet{yoon2025reasoningmodelsbetterexpress} find that RLVR-style reasoning benefits LLMs in calibrating the confidence of their own answers, which seems to contradict our findings at first glance. However, our evaluation suite focuses on predicting human disagreement instead of the models' confidence / uncertainty based on its internal knowledge. The seemingly contradictory results from our work and \citet{yoon2025reasoningmodelsbetterexpress} reflect that calibration and disagreement modeling are orthogonal abilities, while both are essential for responsible decision making. For example, there is one data point where 40\% of human disagree with the majority label (60\%). If a model predicts the majority label with 100\% confidence, it achieves zero calibration error. However, if the confidence score is directly interpreted as a disagreement modeling, it fails to capture any critical disagreement.

Moreover, we find that although scaling LLM size and few-shot steering improve disagreement modeling, these methods are not more effective than a data-centric approach---fine-tuning small LLMs with thousands of human data (\cref{sec:data_importance}). Given the scarcity of repeated human labels, future work may explore how to leverage human data more efficiently.

\section*{Limitations}
This work evaluates the impact of LLM reasoning on disagreement modeling and draws observations with statistical significance tests. Through qualitative analyses, we find that RLVR LLMs tend to assume that all annotators would process the annotation guideline in the same objective way, while RLHF LLM tend to consider annotators' diverse background, although they are prompted with both instructions. However, we fail to draw significant qualitative observations to explain other observations in the paper. For example, why does few-shot steering work for some tasks but not others? Why does scaling in LLM size increase some tasks but not others? These questions are critical to providing concrete guidelines for real-world practice of disagreement modeling. Given our focus on reasoning and the complexity of these question, we leave them for future exploration. 


\section*{Ethics Statement}
\myparagraph{Data Privacy or Bias.} We use publically available datasets (GHC, GoEmotions, and HelpSteer2) which have no data privacy issues or bias against certain demographics. All artifacts we use are under licenses allowing research usage. We also notice no ethical risks associated with this work.
\myparagraph{Reproducibility.} We fully open source our code, prompts, processed datasets, LLM generations, and instructions to reproduce results in \url{https://github.com/EdisonNi-hku/Disagreement_Prediction}.


\bibliography{anthology,custom}

\appendix

\section{Dataset Preparation} \label{appendix:data_preparation}

\begin{figure}[ht]
    \centering
    \includegraphics[width=0.95\columnwidth]{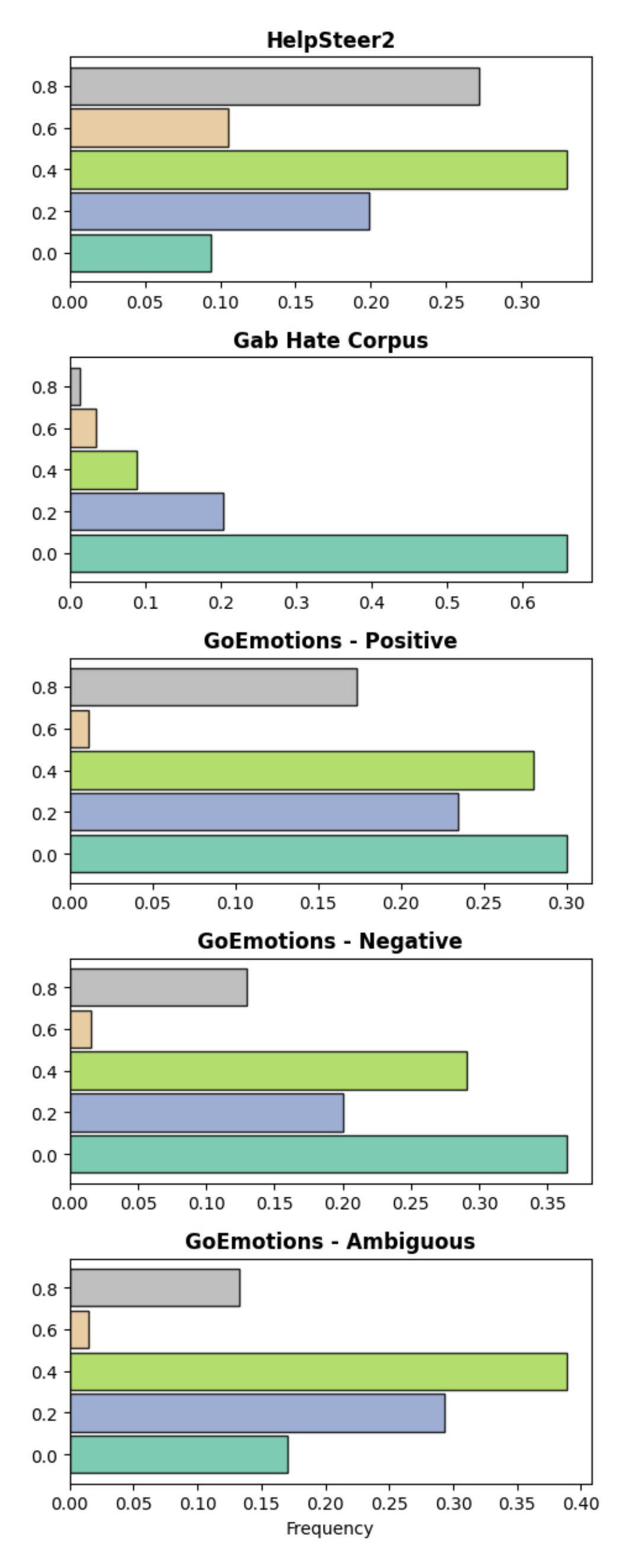}
    \caption{Density bars of the Five Random Sets}
    \label{fig:density_bars}
\end{figure}

For all datasets, we only use the data points with at least 4 annotators for both training and evaluation to ensure annotation quality. Data points with 3 annotations may have one annotator disagree with the others, and the disagreement might be caused by random annotation error (e.g., a wrong click). As shown by \citep{sandri-etal-2023-dont}, 2 annotators making random mistake might be 100 times less likely than 1 annotator doing that.

After this filtering,  we randomly select 2,000 data points from the 3,330 Gab Hate Corpus samples, 2,000 data points from the 20,014 GoEmotions samples, and 1,250 data points from the 2,467 HelpSteer2 samples as training data; and 1K datapoints for \texttt{Random} subsets for testing. The size of training set is strategically picked so that there are enough annotations with high human annotation variance to form the \texttt{HighVar} subsets. HelpSteer2 has a smaller training set because it has less datapoints with at least 4 annotations. Therefore, we shrink its training sets' size to ensure the size of evaluation sets.

The distributions of human annotation $p_d$ of each dataset are presented in \cref{fig:density_bars}.

\section{Distribution Formulas} \label{appendix:distribution_formula}
We probe prediction distributions from large language models (LLMs) using two approaches.

\myparagraph{Sampling-based Distribution.}
We draw $N$ responses $\{y^{(1)}, \dots, y^{(N)}\}$ for the same prompt $p_s$ and estimate the predictive distribution as
\begin{equation}
\hat{p}_{\text{sample}}(y \mid p_s)
\;=\;
\frac{1}{N} \sum_{i=1}^{N} \mathbf{1}\!\left[y^{(i)} = y\right].
\end{equation}
This estimator approximates the model’s implicit predictive distribution via empirical answer frequencies.
The sampling prompt $p_s$ asks the model to produce the most likely human annotation.

\myparagraph{Verbalized Distribution.}
Alternatively, we directly prompt the LLM to report probabilities over the label space $\mathcal{Y}$, yielding
\begin{equation}
\hat{p}_{\text{verbal}}(y \mid p_v) \in \Delta^{|\mathcal{Y}|-1},
\end{equation}
where $\Delta^{|\mathcal{Y}|-1}$ denotes the $(|\mathcal{Y}|-1)$-dimensional probability simplex.
This formulation reflects the model’s self-reported uncertainty.
The verbalization prompt $p_v$ asks the model to estimate the proportion of humans who would predict label $y$.

\section{Prompts} \label{appendix:prompts}
Here we use Gab Hate Corpus prompts to showcase our prompt design. Tasks for other datasets can be found in our Github repository.

One example prompt for Gab Hate Corpus is illustrated bellow, with verbalized distribution and w/o few-shot steering. It is used for RLVR LLMs and for RLHF w/o reasoning, since no reasoning is required in output format. In the instruction part, the model is guided to consider potential disagreement.

{\small
\begin{tcolorbox}[
    breakable,
    colframe=boxborder,    
    colback=boxbg,         
    coltitle=boxtitle,     
    colbacktitle=boxheader,
    fonttitle=\bfseries,   
    boxrule=1pt,           
    width=\columnwidth,  
    title=GHC Prompt Example, 
    enhanced,              
    arc=5pt,               
    left=6pt, right=6pt, top=4pt, bottom=4pt 
]

You are a helpful research assistant with expertise in detecting hate speech from social media posts.\\
\\
Objective:\\
You are tasked with evaluating social media posts to assess the distribution of labels assigned by annotators regarding whether a post constitutes hate speech. The goal is to predict the percentage of annotators who classify the post as hate speech and those who classify it as not hate speech, based on the specific aspects of hate speech it may exhibit.\\
\\
A post contains hate speech if it contains any of the following aspects:\\
- Assaults on Human Dignity: Does the post demean or degrade individuals or groups based on race, ethnicity, gender, religion, sexual orientation, or other protected characteristics?\\
- Calls for Violence: Does the post incite or encourage physical harm or violence against individuals or groups?\\
- Vulgarity and/or Offensive Language: Does the post contain profanity, slurs, or other offensive language that may or may not be directed at individuals or groups?\\
\\
Instruction:\\
- Review the Post: Carefully read the social media post and consider its content in relation to the three aspects of hate speech listed above.\\
- Assess Agreement: Human annotators may have high disagreement if the post contains ambiguous language or context, or if it is likely to be interpreted differently based on subjective views about the severity or nature of hate speech. They may have low disagreement if the post clearly falls into one or more of the hate speech categories or is obviously not hate speech.\\
- Context Sensitivity: Consider the cultural, social, and linguistic context that may influence the interpretation of the post.\\

Output Format:\\%
\relax[Percentage]: a floating-point number indicating the percentages of annotators who classify the post as hate speech, in a squared bracket, e.g., [0.3]. Omit explanations and only give the percentage.

Here is the post: {post}

\end{tcolorbox}
}

For sampling-based distribution, the objective and output format changes to follows, where the LLM is asked to predict the ``most likely'' annotation from human.

{\small
\begin{tcolorbox}[
    breakable, 
    colframe=boxborder,    
    colback=boxbg,         
    coltitle=boxtitle,     
    colbacktitle=boxheader,
    fonttitle=\bfseries,   
    boxrule=1pt,           
    width=\columnwidth,  
    title=Sampling-Based Distribution, 
    enhanced,              
    arc=5pt,               
    left=6pt, right=6pt, top=4pt, bottom=4pt 
]

...\\
Objective:\\
You are tasked with evaluating social media posts to determine whether they contain hate speech or not. The goal is to predict the label most likely assigned by human annotators, classifying each post as either containing hate speech or not, based on the specific aspects of hate speech it expresses.\\
\\
...{task\_description}...\\
\\
Output Format:\\%
\relax[Prediction]: your prediction of how the majority of human annotators will label the post (0 or 1 in squared brackets). [1] if it contains hate speech; [0] if it contains no hate speech.  Omit explanations and only give the percentage.\\
...\\
\end{tcolorbox}
}

When using RLHF LLMs with CoT, we change the output format requirements to:
{\small
\begin{tcolorbox}[
    breakable, 
    colframe=boxborder,    
    colback=boxbg,         
    coltitle=boxtitle,     
    colbacktitle=boxheader,
    fonttitle=\bfseries,   
    boxrule=1pt,           
    width=\columnwidth,  
    title=RLHF CoT, 
    enhanced,              
    arc=5pt,               
    left=6pt, right=6pt, top=4pt, bottom=4pt 
]

...\\
Output Format:\\%
\relax[Reasoning]: your careful reasoning about the content of the post, focusing on the clarity or ambiguity of the post's content concerning aspects of hate speech.\\%
\relax[Prediction]: your prediction of how the majority of human annotators will label the post (0 or 1 in squared brackets). [1] if it contains hate speech; [0] if it contains no hate speech.\\
...
\end{tcolorbox}
}

For few-shot steering, we add the following in-context examples. The few-shot illustrations are carefully picked to avoid biasing the output distribution \citep{turpin2023languagemodelsdontsay}.

{\small
\begin{tcolorbox}[
    breakable, 
    colframe=boxborder,    
    colback=boxbg,         
    coltitle=boxtitle,     
    colbacktitle=boxheader,
    fonttitle=\bfseries,   
    boxrule=1pt,           
    width=\columnwidth,  
    title=GHC Few-Shot Steering, 
    enhanced,              
    arc=5pt,               
    left=6pt, right=6pt, top=4pt, bottom=4pt 
]

...\\
Here are some sample posts:\\

Post: Lots of references to an oligarchy in the shadows who wants to depopulate the planet and sees common folk as insects to be crushed; but who are they? https://www.zerohedge.com/news/2018-08-26/retired-... \#Geopolitics \#Ebola \#Africa \#China | Percetage of postive human labels: 0.25 \\
\newline
Post: The minority that will outbreed you  and make you the minority  | Percetage of postive human labels: 0.50\\\newline
Post: GOTTA REMEMBER...THE LEFT IS INSANELY DESPERATE CAUSE THEY KNOW THEY'RE ON THE PRECIPICE OF GOING TO GITMO FOR TREASON...IT'S ALREADY IN MOTION...SO LET EM SCRAMBLE AND TALK SHIT...THEIR TIME IS LIMITED...LIKE THE DEVIL'S... | Percetage of postive human labels: 1.00 \\\newline
Post: Dozens at Facebook Unite to Challenge Its \u2018Intolerant\u2019 Liberal Culture https://nyti.ms/2NsJ2rJ   It would be nice if it was hundreds, but it's a start. | Percetage of postive human labels: 0.00 \\\newline
Post: I am officially done trying to have peaceful conversation with liberals. They are animals. They deserve to be removed from our republic. Either your believe in America or you are a fucking traitor. Get out of America. You are not welcomed by those of us who love our country. | Percetage of postive human labels: 0.75\\
\end{tcolorbox}
}

\section{Causal Motivation of Our Evaluation Design} \label{app:causal}

To estimate the causal effect of reasoning ($R$) on disagreement modeling ($Y$), it is crucial to account for other experimental factors---such as distribution expression ($X_1$), steering method ($X_2$), and LLM size ($X_3$)---that may influence both $R$ and $Y$. These act as potential confounders.

\myparagraph{Causal Structure.}
The underlying causal graph can be represented as:
\[
X_1, X_2, X_3 \rightarrow R \rightarrow Y,\quad X_1, X_2, X_3 \rightarrow Y
\]
where arrows from $X_i$ to $R$ and $Y$ indicate confounding.

\myparagraph{Backdoor Adjustment.}
To identify the causal effect of $R$ on $Y$, we must block backdoor paths via all $X_i$. This motivates evaluating all combinations so that comparisons between reasoning settings are not confounded by $X_i$.

\myparagraph{Estimand.}
The average causal effect (ACE) of reasoning setting $R$ (vs.\ another reasoning setting $R'$):
\begin{align*}
\mathrm{ACE} = \;
  \mathbb{E}_{x_1, x_2, x_3} \big[\, 
    & Y(r, x_1, x_2, x_3) \\
  -\; & Y(r', x_1, x_2, x_3) \,\big]
\end{align*}
which requires averaging over all settings of $X_1, X_2, X_3$.

\noindent
\textbf{Conclusion.}
By systematically evaluating all factor combinations, we obtain unbiased estimates of the causal effect of reasoning, as detailed by standard causal inference theory \citep{pearl2009causality}.

\section{Inference Details} \label{appendix:hyperparameter}

\myparagraph{LLMs.} We use the following LLMs---RLHF LLMs: \texttt{Llama-3.1-Tulu-3.1-8B}\footnote{\texttt{Llama-3.1-8B-Instruct} from Meta refuse classify hate speeches, so we use Tulu-3.1 which is also based on Llama-3.1-8B}; \texttt{Qwen2.5-14B-Instruct}; \texttt{Qwen2.5-32B-Instruct}; \texttt{Llama-3.3-70B-Instruct}, and \texttt{DeepSeek-V3}. RLVR LLMs: \texttt{DeepSeek-R1-Distill-Llama-8B}; \texttt{DeepSeek-R1-Distill-Qwen-14B}; \texttt{DeepSeek-R1-Distill-Qwen-32B}; \texttt{DeepSeek-R1-Distill-Llama-70B}; and \texttt{DeepSeek-R1}.

\myparagraph{Framework and Hyperparameters.} For 8B to 70B LLMs, we rely on a cluster with 4 GH200 GPUs for local inference. We use vLLM for fast inference. For R1-series RLVR LLMs, we use all official recommended settings, including a temperature of 0.6, and always add <think> at the beginning of assistant message. For RLHF LLMs, we use temperature 0 for verbalized distribution and 0.7 for sampling-based distribution. All other hyperparameters are set to default without restriction on generation length. For the 671B LLMs, we use DeepSeek API with recommended settings. 

\myparagraph{Computational Cost.} The majority of inference cost goes to RLVR LLMs. For the RLVR LLMs of 70B, 32B, 14B, and 8B, the inference costs 100, 40, 20, and 10 GPU hours correspondingly, where the majority is spent on sampling-based distribution which requires sampling 10 times. For RLHF LLMs, especially without CoT, the cost is much less. The RLHF LLMs of 70B, 32B, 14B, and 8B cost 40, 20, 10, 10 GPU hours correspondingly with the cost of CoT and no-CoT settings combined. Note that model loading times are not counted into GPU cost. The API cost of DeepSeek-R1 and DeepSeek-V3 costs roughly 40 USD in total.

\myparagraph{Packages for Evaluation.} Scipy is used to calculate Pearson's Correlations and Wilcoxon Tests.

\section{Fine-Tuning Details}
\label{appendix:ft_detail}
We use Huggingface to fine-tune and evaluate fine-tuned ModernBERT-large and DeBERTa-V3-large. We use a learning rate of 5e-5, a weight decay of 0.01, a batch size of 128, and a epoch number of 5. All other hyperparameters are set to default. 

\section{Results w/o Aggregation} \label{appendix:full_performance}
Here we present the performance of all LLMs with different settings regarding distribution expression, steering, and reasoning, which can be used to calculate all the aggregated results in \cref{sec:result}. Results on \texttt{Random} and \texttt{HighVar} subsets are presented in \cref{tab:rnd_performance_no_agg} and \cref{tab:dis_performance_no_agg}, respectively.

\begin{table*}[h]
\centering
\resizebox{\textwidth}{!}{
\input{figures/tab_appendix_rnd_colored}
}
\caption{Performance on \texttt{Random} (randomly sampled) subsets of all datasets. 
}
\label{tab:rnd_performance_no_agg}
\end{table*}

\begin{table*}[h]
\small
\centering
\resizebox{1.2\columnwidth}{!}{
\input{figures/tab_appendix_dis_colored}
}
\caption{DistAlign Performance on \texttt{HighVar} (high annotation variance) subset of all datasets.
}
\label{tab:dis_performance_no_agg}
\end{table*}

\section{Majority Label Prediction} \label{appendix:f1_analyses}
In \cref{sec:verb_vs_sample}, we observe that sampling-based method achieves better majority label prediction (F1) than verbalized distribution. The prediction of majority labels lies outside the scope of this project, so we analyze those observations in this appendix section to fully reveal the potential of sampling-based methods. We draw the following observations with statistical significance.

\begin{enumerate}
    \item RLVR LLMs outperform RLHF LLMs, with a win rate \colorbox{Green3!15}{$62.50^{**}$\%}.
    \item RLHF w/ CoT outperforms w/o CoT, with a win rate \colorbox{Green3!15}{$62.50^{**}$\%}.
    \item Few-shot steering improves the F1 of GHC with a rate of \colorbox{Green3!20}{$66.67^{**}$\%}, but decrease the HS2, Pos, and Neg where the win rates are \colorbox{Red1!52}{$6.67^{**}$\%}, \colorbox{Red1!20}{$33.33^{**}$\%}, and \colorbox{Red1!28}{$26.67^{**}$\%} correspondingly.
\end{enumerate}

All other trends on F1 do not have statistical significance.

\section{Per-Dataset Results} \label{appendix:perdata}
When comparing RLVR with RLHF LLMs on each dataset, the trends are mostly consistent with \cref{tab:stat_sig} row 2 on \texttt{Random} F1 and \texttt{HighVar} DistAlign. For \texttt{Random} VarCorr and DistAlgin, we further find that following observations with statistical significance: (1) RLVR underperforms RLHF on HS2 \texttt{Random}; and (2) RLVR outperforms RLHF on Pos \texttt{Random}. The trends in \cref{tab:stat_sig} summarizes this observation, as RLVR vs. RLHF has more mixed results on distribution prediction of \texttt{Random} subsets, compared to \texttt{HighVar} subsets.

For CoT vs. w/o CoT on RLHF LLMs, per-dataset comparison shows that on all datasets, CoT either significantly outperforms w/o CoT, or CoT slightly underperforms w/o CoT but without statistical significance.

Furthermore, extending reasoning with RLVR LLMs does not lead to significant change to the performance on all datasets; while verbalized distribution constantly performs significantly better than sampling-based distribution on all datasets.

\section{Qualitative Analysis Details} \label{app:qualitative}

We provide here qualitative analyses of our major quantitative findings on RLVR vs. RLHF models with CoT reasoning.

We sampled 40 datapoints from the GHC dataset---20 with high human disagreement (HighVar) and 20 with low disagreement (distribution of 0 or 1)---focusing on cases where DeepSeek-R1 and V3 made the most divergent predictions. The RLVR LLM tends to adhere more closely to the annotation guidelines. For example, if there is no explicit use of slurs, profanity, or offensive language, it typically assumes that all annotators would agree on a score of 0. Here is one example data point: \textit{Is it not a glorious sight to see the left freak out over the supreme justice retirement announcement? I ask you Morton Kondracke!}

By contrast, the RLHF LLM extrapolates beyond the guidelines and is more likely to consider perspectives from individuals with diverse backgrounds. For instance, R1 does not consider references to “the left” as targeting a protected group, whereas V3 argues that people identifying with this group may find such language offensive. This is notable, given that both models are explicitly prompted to consider diversity of opinion and background. Here is one example data point: \textit{ BREAKING! TECH COMPANIES MEET IN SECRET TO STEAL 2018 ELECTION    Left desperate to win midterms.}

When V3 and R1 disagree, it is more likely (87.5\% of analyzed cases) that V3’s prediction exhibits higher variance than R1’s. This helps explain why RLHF outperforms on HighVar cases, while RLVR performs better on the Random samples, which contains more items with low annotation variance.

In addition, when posts include exclamation marks or words in all capital letters, RLHF models are more likely to assign higher hate speech scores. RLVR models also assign higher scores than human annotators in these cases, but their scores remain lower than those of RLHF models. For instance, a post with the text ``\textit{DEFEND EUROPA}'' is assigned hate speech score of 0.3 by an R1 model and 0.6 by the its RLHF counterpart with CoT, where human label is 0. Moreover, hate speech that does not explicitly use vulgar language is less likely to receive high scores from RLHF models (e.g. ``\textit{So I get accused of being an Islamophobe by a Muzlime on Twitter, I agree with him that I am an Islamofauxb and tell him why and I'm the one who gets suspended?}'' human label: 0.75; RLVR model: 0.7; RLHF model w/ CoT: 0.3). These findings suggest that RLHF models often rely on superficial features when assigning scores.

Furthermore, RLVR models are more likely than RLHF models to assign high hate speech scores to posts related to political issues. Specifically, when posts endorse anti-environmentalism, anti-abortion, patriotism, or general conservative views, RLVR models tend to assign high hate speech scores, whereas human annotators generally agree that such posts are not hate speech.

We conducted similar analyses on the GoEmotions Positive dataset and observed comparable patterns—R1 adheres more strictly to the annotation guidelines, while V3 accounts for a broader range of possible opinions.

\end{document}

%% file: figures/stat_significance.tex
\begin{tabular}{lc@{\hspace{3mm}}c@{\hspace{3mm}}c@{\hspace{3mm}}c}
\toprule
\hspace{4mm} & \bf\tt Random & \bf\tt Random & \bf\tt Random & \bf \tt HighVar \\
& \textbf{VarCorr} & \textbf{DistAlign} & \textbf{F1} & \textbf{DistAlign} \\
\midrule
\multicolumn{4}{l}{\textit{Verbalized > Sampling:}} \\
& \cellcolor{Green3!54} 95.0$^{**}$ & \cellcolor{Green3!51} 92.5$^{**}$ & \cellcolor{Red1!26} 28.3$^{**}$ & \cellcolor{Green3!58} 98.3$^{**}$ \\ 
\multicolumn{4}{l}{\textit{RLVR > RLHF:}} \\
& \cellcolor{Red1!12} 40.0 & \cellcolor{Green3!14} 62.0$^{*}$ & \cellcolor{Red1!17} 36.0$^{**}$  & \cellcolor{Red1!38} 18.0$^{**}$ \\ 
\multicolumn{4}{l}{\textit{RLHF CoT > RLHF w/o CoT :}} \\
& \cellcolor{Green3!17} 64.0$^{**}$ & \cellcolor{Green3!26} 72.0$^{**}$ & \cellcolor{Green3!19} 66.0$^{**}$ & \cellcolor{Green3!24} 70.0$^{**}$ \\ 

\multicolumn{5}{l}{\textit{Extend Reasoning Once > Natural Ending :}} \\
& \cellcolor{Green3!15} 62.5 & \cellcolor{Green3!18} 65.0$^{*}$ & \cellcolor{Red1!3} 47.5 & \cellcolor{Green3!12} 60.0 \\ 

\multicolumn{5}{l}{\textit{Extend Reasoning Twice > Natural Ending :}} \\
& \cellcolor{Green3!12} 60.0 & \cellcolor{Green3!27} 72.5 & \cellcolor{Red1!-0} 50.0 & \cellcolor{Green3!9} 57.5 \\ 
\multicolumn{4}{l}{\textit{w/ > w/o Few-Shot:}} \\
& \cellcolor{Red1!6} 45.3 & \cellcolor{Red1!10} 41.3$^{**}$ & \cellcolor{Red1!23} 30.7$^{**}$ & \cellcolor{Red1!15} 37.3$^{*}$ \\
\midrule
\multicolumn{4}{l}{\textit{HS2 w/ > w/o Few-Shot:}} \\
& \cellcolor{Red1!28} 26.7$^{**}$ & \cellcolor{Red1!60} 0.0$^{**}$ & \cellcolor{Red1!52} 6.7$^{**}$ & \cellcolor{Red1!60} 0.0$^{**}$ \\ 

\multicolumn{4}{l}{\textit{GHC w/ > w/o Few-Shot:}} \\
& \cellcolor{Green3!36} 80.0$^{**}$ & \cellcolor{Green3!36} 80.0$^{**}$ & \cellcolor{Green3!20} 66.7$^{**}$ & \cellcolor{Green3!4} 53.3 \\ 

\multicolumn{4}{l}{\textit{GE-Pos w/ > w/o Few-Shot:}} \\ 
& \cellcolor{Green3!4} 53.3 & \cellcolor{Green3!12} 60.0 & \cellcolor{Red1!20} 33.3$^{**}$ & \cellcolor{Green3!20} 66.7$^{**}$ \\ 

\multicolumn{4}{l}{\textit{GE-Neg w/ > w/o Few-Shot:}} \\
& \cellcolor{Green3!4} 53.3 & \cellcolor{Green3!4} 53.3 & \cellcolor{Red1!28} 26.7$^{**}$ & \cellcolor{Green3!4} 53.3 \\ 

\multicolumn{4}{l}{\textit{GE-Amb w/ > w/o Few-Shot:}} \\
& \cellcolor{Red1!44} 13.3$^{**}$ & \cellcolor{Red1!44} 13.3$^{**}$ & \cellcolor{Red1!36} 20.0 & \cellcolor{Red1!44} 13.3$^{**}$ \\

\midrule
\multicolumn{4}{l}{\textit{Positive > Negative Scaling:}} \\
& \cellcolor{Green3!28} 73.3$^{**}$ & \cellcolor{Green3!24} 70.0$^{**}$ & \cellcolor{Green3!44} 86.7$^{**}$ & \cellcolor{Green3!8} 56.7$^{*}$ \\

\bottomrule
\end{tabular}

%% file: figures/tab_performance_metrics_corrected.tex
\begin{tabular}{llccc|ccc|ccc|ccc|ccc}
\toprule
 & & \multicolumn{3}{c|}{\textbf{HelpSteer2}} & \multicolumn{3}{c|}{\textbf{Gab Hate Corpus}} & \multicolumn{3}{c|}{\textbf{GE-Positive}} & \multicolumn{3}{c|}{\textbf{GE-Negative}} & \multicolumn{3}{c}{\textbf{GE-Ambiguous}} \\
 & & VarCorr$\uparrow$ & DistAlign$\downarrow$ & F1$\uparrow$ & VarCorr$\uparrow$ & DistAlign$\downarrow$ & F1$\uparrow$ & VarCorr$\uparrow$ & DistAlign$\downarrow$ & F1$\uparrow$ & VarCorr$\uparrow$ & DistAlign$\downarrow$ & F1$\uparrow$ & VarCorr$\uparrow$ & DistAlign$\downarrow$ & F1$\uparrow$ \\
\midrule
\rowcolor{gray!20}
 \multicolumn{17}{c}{\textit{Fine-Tuning-Based Methods}} \\
\midrule
\multicolumn{2}{l}{ModernBERT}
 & \cellcolor{Maroon1!0} 0.003 & \cellcolor{Blue1!22} 0.269 & \cellcolor{orange!0} 0.559 & \cellcolor{Maroon1!21} 0.426 & \cellcolor{Blue1!46} 0.141 & \cellcolor{orange!17} 0.368 & \cellcolor{Maroon1!37} 0.277 & \cellcolor{Blue1!52} 0.187 & \cellcolor{orange!38} 0.681 & \cellcolor{Maroon1!51} 0.487 & \cellcolor{Blue1!53} 0.180 & \cellcolor{orange!24} 0.584 & \cellcolor{Maroon1!50} 0.249 & \cellcolor{Blue1!54} 0.198 & \cellcolor{orange!39} 0.528 \\
\multicolumn{2}{l}{DeBERTa-V3}
 & \cellcolor{Maroon1!4} 0.020 & \cellcolor{Blue1!19} 0.272 & \cellcolor{orange!6} 0.578 & \cellcolor{Maroon1!60} 0.554 & \cellcolor{Blue1!60} 0.115 & \cellcolor{orange!47} 0.495 & \cellcolor{Maroon1!60} 0.336 & \cellcolor{Blue1!60} 0.178 & \cellcolor{orange!60} 0.745 & \cellcolor{Maroon1!60} 0.530 & \cellcolor{Blue1!60} 0.168 & \cellcolor{orange!60} 0.670 & \cellcolor{Maroon1!60} 0.289 & \cellcolor{Blue1!60} 0.186 & \cellcolor{orange!60} 0.631 \\

\midrule
\rowcolor{gray!20}
 \multicolumn{17}{c}{\textit{Verbalized Distribution \& \textbf{w/o} Few-shot Steering}} \\
\midrule
\multirow{3}{*}{Avg}
 &  No-CoT  & \cellcolor{Maroon1!37} 0.143 & \cellcolor{Blue1!36} 0.254 & \cellcolor{orange!52} 0.718 & \cellcolor{Maroon1!2} 0.362 & \cellcolor{Blue1!-0} 0.229 & \cellcolor{orange!0} 0.294 & \cellcolor{Maroon1!0} 0.183 & \cellcolor{Blue1!-0} 0.249 & \cellcolor{orange!12} 0.607 & \cellcolor{Maroon1!20} 0.337 & \cellcolor{Blue1!6} 0.265 & \cellcolor{orange!15} 0.561 & \cellcolor{Maroon1!11} 0.096 & \cellcolor{Blue1!16} 0.273 & \cellcolor{orange!22} 0.440 \\
 &  CoT     & \cellcolor{Maroon1!46} 0.177 & \cellcolor{Blue1!40} 0.250 & \cellcolor{orange!39} 0.677 & \cellcolor{Maroon1!2} 0.363 & \cellcolor{Blue1!14} 0.203 & \cellcolor{orange!18} 0.373 & \cellcolor{Maroon1!4} 0.192 & \cellcolor{Blue1!19} 0.226 & \cellcolor{orange!23} 0.638 & \cellcolor{Maroon1!18} 0.329 & \cellcolor{Blue1!16} 0.246 & \cellcolor{orange!18} 0.570 & \cellcolor{Maroon1!16} 0.116 & \cellcolor{Blue1!27} 0.252 & \cellcolor{orange!20} 0.431 \\
 &  R1      & \cellcolor{Maroon1!35} 0.136 & \cellcolor{Blue1!43} 0.247 & \cellcolor{orange!48} 0.705 & \cellcolor{Maroon1!6} 0.374 & \cellcolor{Blue1!27} 0.177 & \cellcolor{orange!23} 0.394 & \cellcolor{Maroon1!21} 0.236 & \cellcolor{Blue1!29} 0.215 & \cellcolor{orange!21} 0.633 & \cellcolor{Maroon1!19} 0.331 & \cellcolor{Blue1!19} 0.242 & \cellcolor{orange!13} 0.556 & \cellcolor{Maroon1!17} 0.121 & \cellcolor{Blue1!24} 0.257 & \cellcolor{orange!12} 0.395 \\
\midrule
\multirow{3}{*}{Best}
 &  No-CoT  & \cellcolor{Maroon1!48} 0.183 & \cellcolor{Blue1!54} 0.236 & \cellcolor{orange!60} 0.741 & \cellcolor{Maroon1!32} 0.461 & \cellcolor{Blue1!37} 0.158 & \cellcolor{orange!19} 0.376 & \cellcolor{Maroon1!23} 0.241 & \cellcolor{Blue1!25} 0.220 & \cellcolor{orange!52} 0.721 & \cellcolor{Maroon1!42} 0.444 & \cellcolor{Blue1!6} 0.265 & \cellcolor{orange!24} 0.583 & \cellcolor{Maroon1!18} 0.126 & \cellcolor{Blue1!25} 0.256 & \cellcolor{orange!43} 0.547 \\
 &  CoT     & \cellcolor{Maroon1!60} 0.230 & \cellcolor{Blue1!59} 0.231 & \cellcolor{orange!51} 0.715 & \cellcolor{Maroon1!13} 0.399 & \cellcolor{Blue1!34} 0.164 & \cellcolor{orange!32} 0.434 & \cellcolor{Maroon1!20} 0.233 & \cellcolor{Blue1!34} 0.209 & \cellcolor{orange!36} 0.675 & \cellcolor{Maroon1!31} 0.389 & \cellcolor{Blue1!16} 0.246 & \cellcolor{orange!23} 0.581 & \cellcolor{Maroon1!33} 0.183 & \cellcolor{Blue1!38} 0.230 & \cellcolor{orange!40} 0.534 \\
 &  R1      & \cellcolor{Maroon1!49} 0.188 & \cellcolor{Blue1!60} 0.230 & \cellcolor{orange!54} 0.722 & \cellcolor{Maroon1!21} 0.426 & \cellcolor{Blue1!43} 0.148 & \cellcolor{orange!39} 0.463 & \cellcolor{Maroon1!36} 0.274 & \cellcolor{Blue1!41} 0.201 & \cellcolor{orange!35} 0.674 & \cellcolor{Maroon1!37} 0.419 & \cellcolor{Blue1!19} 0.241 & \cellcolor{orange!29} 0.596 & \cellcolor{Maroon1!24} 0.147 & \cellcolor{Blue1!36} 0.233 & \cellcolor{orange!26} 0.463 \\
\midrule
\rowcolor{gray!20}
 \multicolumn{17}{c}{\textit{Verbalized Distribution + Few-shot Steering}} \\
\midrule
\multirow{3}{*}{Avg}
 &  No-CoT   & \cellcolor{Maroon1!25} 0.098 & \cellcolor{Blue1!-0} 0.291 & \cellcolor{orange!41} 0.683 & \cellcolor{Maroon1!0} 0.355 & \cellcolor{Blue1!13} 0.205 & \cellcolor{orange!18} 0.372 & \cellcolor{Maroon1!5} 0.197 & \cellcolor{Blue1!8} 0.240 & \cellcolor{orange!0} 0.573 & \cellcolor{Maroon1!0} 0.241 & \cellcolor{Blue1!-0} 0.275 & \cellcolor{orange!0} 0.526 & \cellcolor{Maroon1!0} 0.055 & \cellcolor{Blue1!-0} 0.306 & \cellcolor{orange!24} 0.450 \\
 &  CoT      & \cellcolor{Maroon1!36} 0.139 & \cellcolor{Blue1!12} 0.279 & \cellcolor{orange!42} 0.686 & \cellcolor{Maroon1!8} 0.380 & \cellcolor{Blue1!25} 0.182 & \cellcolor{orange!26} 0.405 & \cellcolor{Maroon1!7} 0.200 & \cellcolor{Blue1!19} 0.226 & \cellcolor{orange!16} 0.619 & \cellcolor{Maroon1!17} 0.321 & \cellcolor{Blue1!14} 0.250 & \cellcolor{orange!17} 0.566 & \cellcolor{Maroon1!11} 0.098 & \cellcolor{Blue1!15} 0.276 & \cellcolor{orange!24} 0.450 \\
 &  R1       & \cellcolor{Maroon1!26} 0.100 & \cellcolor{Blue1!10} 0.281 & \cellcolor{orange!16} 0.608 & \cellcolor{Maroon1!18} 0.416 & \cellcolor{Blue1!37} 0.159 & \cellcolor{orange!23} 0.393 & \cellcolor{Maroon1!21} 0.236 & \cellcolor{Blue1!31} 0.212 & \cellcolor{orange!6} 0.589 & \cellcolor{Maroon1!24} 0.359 & \cellcolor{Blue1!24} 0.233 & \cellcolor{orange!5} 0.538 & \cellcolor{Maroon1!13} 0.107 & \cellcolor{Blue1!13} 0.279 & \cellcolor{orange!0} 0.333 \\

\midrule
\multirow{3}{*}{Best}
 &  No-CoT  & \cellcolor{Maroon1!42} 0.163 & \cellcolor{Blue1!32} 0.258 & \cellcolor{orange!50} 0.710 & \cellcolor{Maroon1!31} 0.459 & \cellcolor{Blue1!46} 0.142 & \cellcolor{orange!60} 0.553 & \cellcolor{Maroon1!26} 0.249 & \cellcolor{Blue1!33} 0.210 & \cellcolor{orange!30} 0.658 & \cellcolor{Maroon1!35} 0.411 & \cellcolor{Blue1!27} 0.226 & \cellcolor{orange!21} 0.576 & \cellcolor{Maroon1!8} 0.088 & \cellcolor{Blue1!19} 0.268 & \cellcolor{orange!40} 0.534 \\
 &  CoT     & \cellcolor{Maroon1!47} 0.182 & \cellcolor{Blue1!25} 0.266 & \cellcolor{orange!44} 0.692 & \cellcolor{Maroon1!24} 0.436 & \cellcolor{Blue1!43} 0.147 & \cellcolor{orange!40} 0.467 & \cellcolor{Maroon1!24} 0.243 & \cellcolor{Blue1!32} 0.211 & \cellcolor{orange!37} 0.680 & \cellcolor{Maroon1!35} 0.409 & \cellcolor{Blue1!31} 0.219 & \cellcolor{orange!22} 0.580 & \cellcolor{Maroon1!21} 0.135 & \cellcolor{Blue1!29} 0.248 & \cellcolor{orange!36} 0.512 \\
 &  R1      & \cellcolor{Maroon1!33} 0.128 & \cellcolor{Blue1!35} 0.255 & \cellcolor{orange!39} 0.678 & \cellcolor{Maroon1!28} 0.449 & \cellcolor{Blue1!49} 0.135 & \cellcolor{orange!35} 0.447 & \cellcolor{Maroon1!27} 0.252 & \cellcolor{Blue1!37} 0.205 & \cellcolor{orange!36} 0.675 & \cellcolor{Maroon1!33} 0.402 & \cellcolor{Blue1!34} 0.214 & \cellcolor{orange!28} 0.593 & \cellcolor{Maroon1!16} 0.118 & \cellcolor{Blue1!19} 0.267 & \cellcolor{orange!21} 0.437 \\

\bottomrule
\end{tabular}

%% file: figures/tab_disagree_subset_perfomance.tex
\begin{tabular}{llccccc}
\toprule
 & & HS2$\downarrow$ & GHC$\downarrow$ & Pos$\downarrow$ & Neg$\downarrow$ & Amb$\downarrow$ \\
\midrule
\rowcolor{gray!20}
 \multicolumn{7}{c}{\textit{Fine-Tuning-Based Methods}} \\
\midrule
\multicolumn{2}{l}{ModernBERT}
  & \cellcolor{Blue1!60} 0.094 & \cellcolor{Blue1!7} 0.246 & \cellcolor{Blue1!60} 0.148 & \cellcolor{Blue1!60} 0.153 & \cellcolor{Blue1!60} 0.138 \\
\multicolumn{2}{l}{DeBERTa-V3}
  & \cellcolor{Blue1!55} 0.109 & \cellcolor{Blue1!-0} 0.256 & \cellcolor{Blue1!53} 0.166 & \cellcolor{Blue1!42} 0.191 & \cellcolor{Blue1!54} 0.153 \\
\midrule
\rowcolor{gray!20}
 \multicolumn{7}{c}{\textit{Verbalized Distribution \& \textbf{w/o} Few-shot Steering}} \\
\midrule
\multirow{3}{*}{Avg}
 &  No-CoT  & \cellcolor{Blue1!4} 0.272 & \cellcolor{Blue1!16} 0.233 & \cellcolor{Blue1!-0} 0.294 & \cellcolor{Blue1!-0} 0.279 & \cellcolor{Blue1!25} 0.223 \\
 &  CoT  & \cellcolor{Blue1!26} 0.202 & \cellcolor{Blue1!34} 0.207 & \cellcolor{Blue1!23} 0.237 & \cellcolor{Blue1!30} 0.217 & \cellcolor{Blue1!37} 0.193 \\
 &  R1  & \cellcolor{Blue1!14} 0.240 & \cellcolor{Blue1!24} 0.222 & \cellcolor{Blue1!14} 0.260 & \cellcolor{Blue1!9} 0.261 & \cellcolor{Blue1!15} 0.246 \\
\midrule
\multirow{3}{*}{Best}
 &  No-CoT  & \cellcolor{Blue1!14} 0.240 & \cellcolor{Blue1!52} 0.182 & \cellcolor{Blue1!18} 0.249 & \cellcolor{Blue1!27} 0.222 & \cellcolor{Blue1!49} 0.165 \\
 &  CoT  & \cellcolor{Blue1!33} 0.180 & \cellcolor{Blue1!60} 0.170 & \cellcolor{Blue1!37} 0.205 & \cellcolor{Blue1!50} 0.173 & \cellcolor{Blue1!53} 0.156 \\
 &  R1  & \cellcolor{Blue1!25} 0.206 & \cellcolor{Blue1!36} 0.204 & \cellcolor{Blue1!32} 0.217 & \cellcolor{Blue1!19} 0.239 & \cellcolor{Blue1!36} 0.195 \\
\midrule
\rowcolor{gray!20}
 \multicolumn{7}{c}{\textit{Verbalized Distribution + Few-shot Steering}} \\
\midrule
\multirow{3}{*}{Avg}
 &  No-CoT  & \cellcolor{Blue1!1} 0.284 & \cellcolor{Blue1!14} 0.236 & \cellcolor{Blue1!25} 0.233 & \cellcolor{Blue1!25} 0.227 & \cellcolor{Blue1!21} 0.233 \\
 &  CoT  & \cellcolor{Blue1!2} 0.279 & \cellcolor{Blue1!31} 0.211 & \cellcolor{Blue1!23} 0.237 & \cellcolor{Blue1!21} 0.234 & \cellcolor{Blue1!22} 0.231 \\
 &  R1  & \cellcolor{Blue1!-0} 0.286 & \cellcolor{Blue1!17} 0.232 & \cellcolor{Blue1!14} 0.260 & \cellcolor{Blue1!9} 0.260 & \cellcolor{Blue1!-0} 0.283 \\
\midrule
\multirow{3}{*}{Best}
 &  No-CoT & \cellcolor{Blue1!22} 0.216 & \cellcolor{Blue1!47} 0.188 & \cellcolor{Blue1!48} 0.178 & \cellcolor{Blue1!57} 0.159 & \cellcolor{Blue1!33} 0.204 \\
 &  CoT    & \cellcolor{Blue1!10} 0.254 & \cellcolor{Blue1!44} 0.193 & \cellcolor{Blue1!38} 0.202 & \cellcolor{Blue1!41} 0.193 & \cellcolor{Blue1!51} 0.159 \\
 &  R1     & \cellcolor{Blue1!11} 0.251 & \cellcolor{Blue1!36} 0.204 & \cellcolor{Blue1!31} 0.218 & \cellcolor{Blue1!24} 0.228 & \cellcolor{Blue1!22} 0.231 \\

\bottomrule
\end{tabular}

%% file: figures/tab_scaling.tex
\begin{tabular}{l@{\hspace{2mm}}c@{\hspace{2mm}}c@{\hspace{2mm}}c@{\hspace{2mm}}c|c@{\hspace{2mm}}c@{\hspace{2mm}}c@{\hspace{2mm}}c|c@{\hspace{2mm}}c@{\hspace{2mm}}c@{\hspace{2mm}}c|c@{\hspace{2mm}}c@{\hspace{2mm}}c@{\hspace{2mm}}c|c@{\hspace{2mm}}c@{\hspace{2mm}}c@{\hspace{2mm}}c}
\toprule
  & \multicolumn{3}{c}{\textbf{HS2 \texttt{Random}}} & \textbf{\texttt{HighVar}} & \multicolumn{3}{c}{\textbf{GHC \texttt{Random}}} & \textbf{\texttt{HighVar}} & \multicolumn{3}{c}{\textbf{Pos \texttt{Random}}} & \textbf{\texttt{HighVar}} & \multicolumn{3}{c}{\textbf{Neg \texttt{Random}}} & \textbf{\texttt{HighVar}} & \multicolumn{3}{c}{\textbf{Amb \texttt{Random}}} & \textbf{\texttt{HighVar}} \\
 & VarCorr & DistAlign & F1 & DistAlgin & VarCorr & DistAlign & F1 & DistAlgin & VarCorr & DistAlign & F1 & DistAlgin & VarCorr & DistAlign & F1 & DistAlgin & VarCorr & DistAlign & F1 & DistAlgin \\
\midrule
\rowcolor{gray!20}
 \multicolumn{21}{c}{\textit{Verbalized Distribution but \textbf{w/o} Few-shot Steering}} \\
\midrule
No-CoT  & \cellcolor{Green3!42} 0.702 & \cellcolor{Green3!42} 0.703 & \cellcolor{Green3!57} 0.945 & \cellcolor{Red1!2} -0.037 & \cellcolor{Red1!21} -0.345 & \cellcolor{Red1!3} -0.049 & \cellcolor{Green3!17} 0.277 & \cellcolor{Green3!43} 0.722 & \cellcolor{Green3!34} 0.568 & \cellcolor{Green3!35} 0.586 & \cellcolor{Green3!50} 0.825 & \cellcolor{Green3!41} 0.690 & \cellcolor{Red1!24} -0.402 & \cellcolor{Red1!12} -0.197 & \cellcolor{Green3!32} 0.539 & \cellcolor{Green3!12} 0.196 & \cellcolor{Green3!49} 0.818 & \cellcolor{Green3!13} 0.224 & \cellcolor{Green3!26} 0.428 & \cellcolor{Red1!3} -0.046 \\
CoT     & \cellcolor{Green3!55} 0.913 & \cellcolor{Green3!44} 0.738 & \cellcolor{Green3!27} 0.447 & \cellcolor{Red1!6} -0.097 & \cellcolor{Green3!26} 0.441 & \cellcolor{Green3!29} 0.485 & \cellcolor{Green3!48} 0.799 & \cellcolor{Green3!16} 0.261 & \cellcolor{Green3!47} 0.786 & \cellcolor{Green3!36} 0.593 & \cellcolor{Green3!35} 0.582 & \cellcolor{Green3!16} 0.260 & \cellcolor{Red1!18} -0.303 & \cellcolor{Red1!17} -0.280 & \cellcolor{Green3!41} 0.686 & \cellcolor{Red1!6} -0.096 & \cellcolor{Green3!54} 0.899 & \cellcolor{Green3!51} 0.854 & \cellcolor{Green3!20} 0.329 & \cellcolor{Green3!8} 0.138 \\
R1      & \cellcolor{Green3!51} 0.852 & \cellcolor{Green3!47} 0.790 & \cellcolor{Green3!44} 0.726 & \cellcolor{Red1!40} -0.668 & \cellcolor{Green3!5} 0.083 & \cellcolor{Red1!24} -0.400 & \cellcolor{Green3!38} 0.628 & \cellcolor{Green3!52} 0.862 & \cellcolor{Red1!4} -0.059 & \cellcolor{Green3!36} 0.598 & \cellcolor{Green3!28} 0.470 & \cellcolor{Green3!51} 0.853 & \cellcolor{Red1!42} -0.700 & \cellcolor{Red1!20} -0.333 & \cellcolor{Green3!18} 0.306 & \cellcolor{Green3!52} 0.873 & \cellcolor{Green3!31} 0.518 & \cellcolor{Green3!56} 0.934 & \cellcolor{Green3!39} 0.657 & \cellcolor{Green3!40} 0.667 \\
\midrule
\rowcolor{gray!20}
 \multicolumn{21}{c}{\textit{Verbalized Distribution + Few-shot Steering}} \\
\midrule
No-CoT  & \cellcolor{Green3!54} 0.906 & \cellcolor{Green3!48} 0.804 & \cellcolor{Green3!30} 0.507 & \cellcolor{Green3!24} 0.399 & \cellcolor{Green3!16} 0.275 & \cellcolor{Green3!18} 0.298 & \cellcolor{Green3!14} 0.240 & \cellcolor{Green3!10} 0.175 & \cellcolor{Green3!35} 0.578 & \cellcolor{Green3!36} 0.593 & \cellcolor{Green3!47} 0.778 & \cellcolor{Red1!17} -0.289 & \cellcolor{Red1!10} -0.167 & \cellcolor{Red1!14} -0.235 & \cellcolor{Green3!2} 0.030 & \cellcolor{Red1!49} -0.819 & \cellcolor{Green3!1} 0.014 & \cellcolor{Green3!1} 0.023 & \cellcolor{Green3!35} 0.584 & \cellcolor{Green3!10} 0.172 \\
CoT     & \cellcolor{Green3!42} 0.692 & \cellcolor{Green3!15} 0.252 & \cellcolor{Red1!13} -0.209 & \cellcolor{Red1!14} -0.230 & \cellcolor{Green3!27} 0.457 & \cellcolor{Green3!28} 0.463 & \cellcolor{Green3!35} 0.587 & \cellcolor{Red1!23} -0.379 & \cellcolor{Green3!30} 0.503 & \cellcolor{Green3!26} 0.428 & \cellcolor{Green3!47} 0.777 & \cellcolor{Red1!3} -0.047 & \cellcolor{Red1!10} -0.170 & \cellcolor{Red1!27} -0.455 & \cellcolor{Green3!18} 0.299 & \cellcolor{Red1!36} -0.604 & \cellcolor{Green3!30} 0.504 & \cellcolor{Green3!20} 0.327 & \cellcolor{Green3!27} 0.457 & \cellcolor{Red1!6} -0.105 \\
R1      & \cellcolor{Green3!39} 0.653 & \cellcolor{Red1!6} -0.104 & \cellcolor{Red1!49} -0.811 & \cellcolor{Red1!29} -0.488 & \cellcolor{Green3!9} 0.151 & \cellcolor{Green3!3} 0.056 & \cellcolor{Green3!32} 0.539 & \cellcolor{Green3!40} 0.671 & \cellcolor{Green3!38} 0.639 & \cellcolor{Green3!42} 0.700 & \cellcolor{Red1!18} -0.299 & \cellcolor{Green3!47} 0.789 & \cellcolor{Red1!43} -0.714 & \cellcolor{Red1!34} -0.570 & \cellcolor{Red1!9} -0.152 & \cellcolor{Green3!48} 0.792 & \cellcolor{Green3!27} 0.449 & \cellcolor{Green3!12} 0.204 & \cellcolor{Green3!52} 0.862 & \cellcolor{Green3!30} 0.504 \\

 \bottomrule
\end{tabular}

%% file: figures/tab_appendix_rnd_colored.tex
\begin{tabular}{llc@{\hspace{2mm}}c@{\hspace{2mm}}c|c@{\hspace{2mm}}c@{\hspace{2mm}}c|c@{\hspace{2mm}}c@{\hspace{2mm}}c|c@{\hspace{2mm}}c@{\hspace{2mm}}c|c@{\hspace{2mm}}c@{\hspace{2mm}}c}
\toprule
 & & \multicolumn{3}{c|}{\textbf{HelpSteer2}} & \multicolumn{3}{c|}{\textbf{Gab Hate Corpus}} & \multicolumn{3}{c|}{\textbf{GE-Positive}} & \multicolumn{3}{c|}{\textbf{GE-Negative}} & \multicolumn{3}{c}{\textbf{GE-Ambiguous}} \\
 & & VarCorr$\uparrow$ & DistAlign$\downarrow$ & F1$\uparrow$ & VarCorr$\uparrow$ & DistAlign$\downarrow$ & F1$\uparrow$ & VarCorr$\uparrow$ & DistAlign$\downarrow$ & F1$\uparrow$ & VarCorr$\uparrow$ & DistAlign$\downarrow$ & F1$\uparrow$ & VarCorr$\uparrow$ & DistAlign$\downarrow$ & F1$\uparrow$ \\

\midrule
\rowcolor{gray!20}
 \multicolumn{17}{c}{\textit{Verbalized Distribution \& \textbf{w/o} Few-shot Steering}} \\
\midrule
\multirow{3}{*}{Llama-8B}
 &  No-CoT   & \cellcolor{Maroon1!11} 0.043 & \cellcolor{Blue1!47} 0.277 & \cellcolor{orange!48} 0.699 & \cellcolor{Maroon1!37} 0.283 & \cellcolor{Blue1!20} 0.290 & \cellcolor{orange!12} 0.225 & \cellcolor{Maroon1!24} 0.109 & \cellcolor{Blue1!6} 0.357 & \cellcolor{orange!18} 0.504 & \cellcolor{Maroon1!36} 0.282 & \cellcolor{Blue1!29} 0.294 & \cellcolor{orange!26} 0.517 & \cellcolor{Maroon1!14} 0.045 & \cellcolor{Blue1!41} 0.309 & \cellcolor{orange!51} 0.499 \\
 &  CoT      & \cellcolor{Maroon1!33} 0.127 & \cellcolor{Blue1!48} 0.273 & \cellcolor{orange!48} 0.699 & \cellcolor{Maroon1!34} 0.262 & \cellcolor{Blue1!26} 0.265 & \cellcolor{orange!18} 0.270 & \cellcolor{Maroon1!26} 0.121 & \cellcolor{Blue1!36} 0.269 & \cellcolor{orange!43} 0.631 & \cellcolor{Maroon1!32} 0.256 & \cellcolor{Blue1!37} 0.269 & \cellcolor{orange!44} 0.566 & \cellcolor{Maroon1!28} 0.089 & \cellcolor{Blue1!50} 0.273 & \cellcolor{orange!54} 0.514 \\
 &  R1       & \cellcolor{Maroon1!13} 0.053 & \cellcolor{Blue1!46} 0.281 & \cellcolor{orange!47} 0.695 & \cellcolor{Maroon1!39} 0.298 & \cellcolor{Blue1!45} 0.194 & \cellcolor{orange!12} 0.230 & \cellcolor{Maroon1!41} 0.186 & \cellcolor{Blue1!46} 0.240 & \cellcolor{orange!27} 0.547 & \cellcolor{Maroon1!39} 0.301 & \cellcolor{Blue1!36} 0.273 & \cellcolor{orange!2} 0.456 & \cellcolor{Maroon1!44} 0.136 & \cellcolor{Blue1!51} 0.268 & \cellcolor{orange!34} 0.408 \\
\midrule
\multirow{3}{*}{Qwen-14B}
 &  No-CoT   & \cellcolor{Maroon1!38} 0.147 & \cellcolor{Blue1!54} 0.251 & \cellcolor{orange!51} 0.713 & \cellcolor{Maroon1!58} 0.442 & \cellcolor{Blue1!41} 0.206 & \cellcolor{orange!22} 0.294 & \cellcolor{Maroon1!38} 0.175 & \cellcolor{Blue1!51} 0.228 & \cellcolor{orange!44} 0.637 & \cellcolor{Maroon1!45} 0.344 & \cellcolor{Blue1!34} 0.280 & \cellcolor{orange!41} 0.558 & \cellcolor{Maroon1!26} 0.083 & \cellcolor{Blue1!51} 0.265 & \cellcolor{orange!31} 0.392 \\
 &  CoT      & \cellcolor{Maroon1!34} 0.132 & \cellcolor{Blue1!53} 0.256 & \cellcolor{orange!22} 0.566 & \cellcolor{Maroon1!52} 0.399 & \cellcolor{Blue1!45} 0.194 & \cellcolor{orange!33} 0.372 & \cellcolor{Maroon1!42} 0.194 & \cellcolor{Blue1!53} 0.222 & \cellcolor{orange!46} 0.647 & \cellcolor{Maroon1!50} 0.374 & \cellcolor{Blue1!47} 0.239 & \cellcolor{orange!47} 0.573 & \cellcolor{Maroon1!21} 0.068 & \cellcolor{Blue1!51} 0.266 & \cellcolor{orange!31} 0.392 \\
 &  R1       & \cellcolor{Maroon1!28} 0.109 & \cellcolor{Blue1!54} 0.252 & \cellcolor{orange!43} 0.675 & \cellcolor{Maroon1!55} 0.426 & \cellcolor{Blue1!55} 0.153 & \cellcolor{orange!37} 0.400 & \cellcolor{Maroon1!56} 0.256 & \cellcolor{Blue1!55} 0.214 & \cellcolor{orange!50} 0.670 & \cellcolor{Maroon1!56} 0.419 & \cellcolor{Blue1!55} 0.215 & \cellcolor{orange!56} 0.596 & \cellcolor{Maroon1!24} 0.076 & \cellcolor{Blue1!51} 0.268 & \cellcolor{orange!22} 0.339 \\
\midrule
\multirow{3}{*}{Qwen-32B}
 &  No-CoT   & \cellcolor{Maroon1!45} 0.172 & \cellcolor{Blue1!56} 0.245 & \cellcolor{orange!52} 0.721 & \cellcolor{Maroon1!60} 0.461 & \cellcolor{Blue1!54} 0.158 & \cellcolor{orange!34} 0.376 & \cellcolor{Maroon1!43} 0.195 & \cellcolor{Blue1!53} 0.220 & \cellcolor{orange!28} 0.552 & \cellcolor{Maroon1!60} 0.444 & \cellcolor{Blue1!60} 0.198 & \cellcolor{orange!51} 0.583 & \cellcolor{Maroon1!33} 0.102 & \cellcolor{Blue1!54} 0.256 & \cellcolor{orange!9} 0.273 \\
 &  CoT      & \cellcolor{Maroon1!50} 0.193 & \cellcolor{Blue1!59} 0.234 & \cellcolor{orange!50} 0.706 & \cellcolor{Maroon1!52} 0.398 & \cellcolor{Blue1!52} 0.164 & \cellcolor{orange!37} 0.400 & \cellcolor{Maroon1!46} 0.210 & \cellcolor{Blue1!55} 0.214 & \cellcolor{orange!36} 0.594 & \cellcolor{Maroon1!52} 0.389 & \cellcolor{Blue1!54} 0.216 & \cellcolor{orange!43} 0.562 & \cellcolor{Maroon1!27} 0.084 & \cellcolor{Blue1!53} 0.257 & \cellcolor{orange!9} 0.270 \\
 &  R1       & \cellcolor{Maroon1!39} 0.151 & \cellcolor{Blue1!56} 0.243 & \cellcolor{orange!51} 0.713 & \cellcolor{Maroon1!55} 0.425 & \cellcolor{Blue1!57} 0.148 & \cellcolor{orange!47} 0.463 & \cellcolor{Maroon1!57} 0.262 & \cellcolor{Blue1!57} 0.209 & \cellcolor{orange!42} 0.625 & \cellcolor{Maroon1!53} 0.398 & \cellcolor{Blue1!56} 0.212 & \cellcolor{orange!50} 0.581 & \cellcolor{Maroon1!40} 0.123 & \cellcolor{Blue1!50} 0.269 & \cellcolor{orange!20} 0.330 \\
\midrule
\multirow{3}{*}{Llama-70B}
 &  No-CoT  & \cellcolor{Maroon1!44} 0.171 & \cellcolor{Blue1!51} 0.263 & \cellcolor{orange!52} 0.717 & \cellcolor{Maroon1!44} 0.337 & \cellcolor{Blue1!33} 0.238 & \cellcolor{orange!19} 0.274 & \cellcolor{Maroon1!53} 0.241 & \cellcolor{Blue1!53} 0.221 & \cellcolor{orange!41} 0.620 & \cellcolor{Maroon1!55} 0.409 & \cellcolor{Blue1!45} 0.245 & \cellcolor{orange!49} 0.579 & \cellcolor{Maroon1!41} 0.126 & \cellcolor{Blue1!53} 0.258 & \cellcolor{orange!49} 0.487 \\
 &  CoT     & \cellcolor{Maroon1!53} 0.205 & \cellcolor{Blue1!53} 0.257 & \cellcolor{orange!48} 0.697 & \cellcolor{Maroon1!49} 0.376 & \cellcolor{Blue1!41} 0.208 & \cellcolor{orange!36} 0.389 & \cellcolor{Maroon1!44} 0.202 & \cellcolor{Blue1!57} 0.209 & \cellcolor{orange!45} 0.644 & \cellcolor{Maroon1!50} 0.379 & \cellcolor{Blue1!49} 0.234 & \cellcolor{orange!45} 0.567 & \cellcolor{Maroon1!51} 0.155 & \cellcolor{Blue1!60} 0.230 & \cellcolor{orange!42} 0.448 \\
 &  R1      & \cellcolor{Maroon1!47} 0.180 & \cellcolor{Blue1!60} 0.230 & \cellcolor{orange!53} 0.722 & \cellcolor{Maroon1!46} 0.351 & \cellcolor{Blue1!45} 0.193 & \cellcolor{orange!42} 0.428 & \cellcolor{Maroon1!60} 0.274 & \cellcolor{Blue1!60} 0.201 & \cellcolor{orange!51} 0.674 & \cellcolor{Maroon1!43} 0.332 & \cellcolor{Blue1!49} 0.234 & \cellcolor{orange!55} 0.595 & \cellcolor{Maroon1!41} 0.125 & \cellcolor{Blue1!56} 0.247 & \cellcolor{orange!40} 0.436 \\
\midrule
\multirow{3}{*}{Deepseek}
 &  V3-no-CoT         & \cellcolor{Maroon1!48} 0.183 & \cellcolor{Blue1!58} 0.236 & \cellcolor{orange!56} 0.741 & \cellcolor{Maroon1!37} 0.288 & \cellcolor{Blue1!29} 0.254 & \cellcolor{orange!23} 0.302 & \cellcolor{Maroon1!42} 0.194 & \cellcolor{Blue1!53} 0.220 & \cellcolor{orange!60} 0.721 & \cellcolor{Maroon1!25} 0.208 & \cellcolor{Blue1!25} 0.307 & \cellcolor{orange!45} 0.568 & \cellcolor{Maroon1!40} 0.123 & \cellcolor{Blue1!48} 0.280 & \cellcolor{orange!60} 0.547 \\
 &  V3-CoT            & \cellcolor{Maroon1!60} 0.230 & \cellcolor{Blue1!60} 0.231 & \cellcolor{orange!51} 0.715 & \cellcolor{Maroon1!49} 0.381 & \cellcolor{Blue1!47} 0.186 & \cellcolor{orange!42} 0.434 & \cellcolor{Maroon1!51} 0.233 & \cellcolor{Blue1!55} 0.216 & \cellcolor{orange!51} 0.675 & \cellcolor{Maroon1!30} 0.246 & \cellcolor{Blue1!36} 0.273 & \cellcolor{orange!50} 0.581 & \cellcolor{Maroon1!60} 0.183 & \cellcolor{Blue1!59} 0.234 & \cellcolor{orange!58} 0.534 \\
 &  R1                & \cellcolor{Maroon1!49} 0.188 & \cellcolor{Blue1!60} 0.231 & \cellcolor{orange!52} 0.721 & \cellcolor{Maroon1!48} 0.370 & \cellcolor{Blue1!44} 0.196 & \cellcolor{orange!44} 0.447 & \cellcolor{Maroon1!45} 0.204 & \cellcolor{Blue1!57} 0.209 & \cellcolor{orange!46} 0.649 & \cellcolor{Maroon1!24} 0.206 & \cellcolor{Blue1!36} 0.274 & \cellcolor{orange!39} 0.552 & \cellcolor{Maroon1!48} 0.147 & \cellcolor{Blue1!59} 0.233 & \cellcolor{orange!44} 0.463 \\

\midrule
\rowcolor{gray!20}
 \multicolumn{17}{c}{\textit{Verbalized Distribution + Few-shot Steering}} \\
\midrule
\multirow{3}{*}{Llama-8B}
 &  No-CoT  & \cellcolor{Maroon1!12} 0.049 & \cellcolor{Blue1!43} 0.293 & \cellcolor{orange!40} 0.658 & \cellcolor{Maroon1!14} 0.111 & \cellcolor{Blue1!-0} 0.365 & \cellcolor{orange!0} 0.147 & \cellcolor{Maroon1!15} 0.070 & \cellcolor{Blue1!17} 0.325 & \cellcolor{orange!0} 0.409 & \cellcolor{Maroon1!1} 0.052 & \cellcolor{Blue1!15} 0.340 & \cellcolor{orange!0} 0.450 & \cellcolor{Maroon1!0} 0.005 & \cellcolor{Blue1!31} 0.347 & \cellcolor{orange!49} 0.489 \\
 &  CoT     & \cellcolor{Maroon1!17} 0.067 & \cellcolor{Blue1!41} 0.297 & \cellcolor{orange!47} 0.692 & \cellcolor{Maroon1!28} 0.215 & \cellcolor{Blue1!22} 0.282 & \cellcolor{orange!12} 0.230 & \cellcolor{Maroon1!31} 0.142 & \cellcolor{Blue1!41} 0.255 & \cellcolor{orange!23} 0.526 & \cellcolor{Maroon1!23} 0.197 & \cellcolor{Blue1!35} 0.276 & \cellcolor{orange!34} 0.540 & \cellcolor{Maroon1!40} 0.123 & \cellcolor{Blue1!51} 0.267 & \cellcolor{orange!50} 0.494 \\
 &  R1      & \cellcolor{Maroon1!16} 0.065 & \cellcolor{Blue1!41} 0.297 & \cellcolor{orange!44} 0.676 & \cellcolor{Maroon1!46} 0.353 & \cellcolor{Blue1!47} 0.186 & \cellcolor{orange!16} 0.258 & \cellcolor{Maroon1!51} 0.234 & \cellcolor{Blue1!52} 0.224 & \cellcolor{orange!26} 0.546 & \cellcolor{Maroon1!46} 0.352 & \cellcolor{Blue1!45} 0.245 & \cellcolor{orange!2} 0.456 & \cellcolor{Maroon1!27} 0.086 & \cellcolor{Blue1!48} 0.279 & \cellcolor{orange!13} 0.290 \\
\midrule
\multirow{3}{*}{Qwen-14B}
 &  No-CoT  & \cellcolor{Maroon1!22} 0.086 & \cellcolor{Blue1!36} 0.317 & \cellcolor{orange!50} 0.710 & \cellcolor{Maroon1!60} 0.459 & \cellcolor{Blue1!58} 0.142 & \cellcolor{orange!60} 0.553 & \cellcolor{Maroon1!45} 0.207 & \cellcolor{Blue1!52} 0.224 & \cellcolor{orange!34} 0.584 & \cellcolor{Maroon1!49} 0.371 & \cellcolor{Blue1!51} 0.226 & \cellcolor{orange!41} 0.557 & \cellcolor{Maroon1!25} 0.079 & \cellcolor{Blue1!46} 0.289 & \cellcolor{orange!28} 0.375 \\
 &  CoT     & \cellcolor{Maroon1!36} 0.139 & \cellcolor{Blue1!50} 0.267 & \cellcolor{orange!45} 0.685 & \cellcolor{Maroon1!56} 0.428 & \cellcolor{Blue1!57} 0.147 & \cellcolor{orange!47} 0.467 & \cellcolor{Maroon1!45} 0.205 & \cellcolor{Blue1!51} 0.226 & \cellcolor{orange!44} 0.639 & \cellcolor{Maroon1!51} 0.387 & \cellcolor{Blue1!52} 0.224 & \cellcolor{orange!50} 0.580 & \cellcolor{Maroon1!8} 0.029 & \cellcolor{Blue1!44} 0.296 & \cellcolor{orange!30} 0.386 \\
 &  R1      & \cellcolor{Maroon1!29} 0.114 & \cellcolor{Blue1!53} 0.255 & \cellcolor{orange!43} 0.674 & \cellcolor{Maroon1!58} 0.442 & \cellcolor{Blue1!60} 0.135 & \cellcolor{orange!44} 0.444 & \cellcolor{Maroon1!47} 0.216 & \cellcolor{Blue1!55} 0.214 & \cellcolor{orange!38} 0.608 & \cellcolor{Maroon1!54} 0.402 & \cellcolor{Blue1!55} 0.214 & \cellcolor{orange!55} 0.593 & \cellcolor{Maroon1!34} 0.105 & \cellcolor{Blue1!51} 0.267 & \cellcolor{orange!2} 0.234 \\
\midrule
\multirow{3}{*}{Qwen-32B}
 &  No-CoT  & \cellcolor{Maroon1!28} 0.108 & \cellcolor{Blue1!43} 0.290 & \cellcolor{orange!39} 0.655 & \cellcolor{Maroon1!56} 0.434 & \cellcolor{Blue1!57} 0.145 & \cellcolor{orange!35} 0.387 & \cellcolor{Maroon1!54} 0.249 & \cellcolor{Blue1!57} 0.210 & \cellcolor{orange!33} 0.582 & \cellcolor{Maroon1!37} 0.288 & \cellcolor{Blue1!46} 0.241 & \cellcolor{orange!40} 0.555 & \cellcolor{Maroon1!28} 0.088 & \cellcolor{Blue1!51} 0.268 & \cellcolor{orange!30} 0.383 \\
 &  CoT     & \cellcolor{Maroon1!37} 0.144 & \cellcolor{Blue1!50} 0.266 & \cellcolor{orange!44} 0.680 & \cellcolor{Maroon1!57} 0.436 & \cellcolor{Blue1!55} 0.154 & \cellcolor{orange!37} 0.397 & \cellcolor{Maroon1!45} 0.205 & \cellcolor{Blue1!56} 0.213 & \cellcolor{orange!35} 0.591 & \cellcolor{Maroon1!53} 0.394 & \cellcolor{Blue1!50} 0.230 & \cellcolor{orange!45} 0.567 & \cellcolor{Maroon1!23} 0.072 & \cellcolor{Blue1!42} 0.302 & \cellcolor{orange!27} 0.368 \\
 &  R1      & \cellcolor{Maroon1!17} 0.066 & \cellcolor{Blue1!41} 0.298 & \cellcolor{orange!20} 0.558 & \cellcolor{Maroon1!58} 0.449 & \cellcolor{Blue1!56} 0.149 & \cellcolor{orange!35} 0.386 & \cellcolor{Maroon1!54} 0.247 & \cellcolor{Blue1!59} 0.205 & \cellcolor{orange!39} 0.610 & \cellcolor{Maroon1!48} 0.365 & \cellcolor{Blue1!52} 0.223 & \cellcolor{orange!46} 0.570 & \cellcolor{Maroon1!38} 0.118 & \cellcolor{Blue1!41} 0.306 & \cellcolor{orange!13} 0.291 \\
\midrule
\multirow{3}{*}{Llama-70B}
 &  No-CoT  & \cellcolor{Maroon1!21} 0.083 & \cellcolor{Blue1!41} 0.299 & \cellcolor{orange!45} 0.684 & \cellcolor{Maroon1!56} 0.431 & \cellcolor{Blue1!52} 0.166 & \cellcolor{orange!34} 0.378 & \cellcolor{Maroon1!50} 0.229 & \cellcolor{Blue1!51} 0.227 & \cellcolor{orange!43} 0.633 & \cellcolor{Maroon1!55} 0.411 & \cellcolor{Blue1!48} 0.236 & \cellcolor{orange!48} 0.576 & \cellcolor{Maroon1!26} 0.083 & \cellcolor{Blue1!40} 0.310 & \cellcolor{orange!46} 0.471 \\
 &  CoT     & \cellcolor{Maroon1!47} 0.182 & \cellcolor{Blue1!41} 0.297 & \cellcolor{orange!46} 0.687 & \cellcolor{Maroon1!54} 0.413 & \cellcolor{Blue1!52} 0.164 & \cellcolor{orange!47} 0.467 & \cellcolor{Maroon1!53} 0.243 & \cellcolor{Blue1!57} 0.211 & \cellcolor{orange!48} 0.656 & \cellcolor{Maroon1!55} 0.409 & \cellcolor{Blue1!53} 0.219 & \cellcolor{orange!48} 0.576 & \cellcolor{Maroon1!43} 0.132 & \cellcolor{Blue1!56} 0.248 & \cellcolor{orange!49} 0.490 \\
 &  R1      & \cellcolor{Maroon1!33} 0.127 & \cellcolor{Blue1!51} 0.261 & \cellcolor{orange!44} 0.678 & \cellcolor{Maroon1!56} 0.433 & \cellcolor{Blue1!53} 0.161 & \cellcolor{orange!44} 0.447 & \cellcolor{Maroon1!51} 0.231 & \cellcolor{Blue1!57} 0.211 & \cellcolor{orange!51} 0.675 & \cellcolor{Maroon1!46} 0.352 & \cellcolor{Blue1!50} 0.229 & \cellcolor{orange!54} 0.592 & \cellcolor{Maroon1!38} 0.118 & \cellcolor{Blue1!49} 0.274 & \cellcolor{orange!35} 0.411 \\
\midrule
\multirow{3}{*}{Deepseek}
 &  V3-no-CoT  & \cellcolor{Maroon1!42} 0.163 & \cellcolor{Blue1!52} 0.258 & \cellcolor{orange!50} 0.710 & \cellcolor{Maroon1!45} 0.343 & \cellcolor{Blue1!41} 0.208 & \cellcolor{orange!37} 0.396 & \cellcolor{Maroon1!50} 0.229 & \cellcolor{Blue1!56} 0.212 & \cellcolor{orange!48} 0.658 & \cellcolor{Maroon1!6} 0.085 & \cellcolor{Blue1!18} 0.331 & \cellcolor{orange!15} 0.490 & \cellcolor{Maroon1!8} 0.028 & \cellcolor{Blue1!39} 0.317 & \cellcolor{orange!58} 0.534 \\
 &  V3-CoT     & \cellcolor{Maroon1!43} 0.164 & \cellcolor{Blue1!49} 0.271 & \cellcolor{orange!46} 0.686 & \cellcolor{Maroon1!53} 0.406 & \cellcolor{Blue1!52} 0.164 & \cellcolor{orange!47} 0.462 & \cellcolor{Maroon1!45} 0.206 & \cellcolor{Blue1!51} 0.226 & \cellcolor{orange!52} 0.680 & \cellcolor{Maroon1!26} 0.220 & \cellcolor{Blue1!27} 0.300 & \cellcolor{orange!44} 0.566 & \cellcolor{Maroon1!44} 0.135 & \cellcolor{Blue1!51} 0.268 & \cellcolor{orange!54} 0.512 \\
 &  R1         & \cellcolor{Maroon1!33} 0.128 & \cellcolor{Blue1!43} 0.291 & \cellcolor{orange!0} 0.455 & \cellcolor{Maroon1!52} 0.403 & \cellcolor{Blue1!53} 0.162 & \cellcolor{orange!42} 0.429 & \cellcolor{Maroon1!55} 0.252 & \cellcolor{Blue1!58} 0.206 & \cellcolor{orange!19} 0.509 & \cellcolor{Maroon1!42} 0.322 & \cellcolor{Blue1!41} 0.257 & \cellcolor{orange!11} 0.479 & \cellcolor{Maroon1!35} 0.107 & \cellcolor{Blue1!50} 0.270 & \cellcolor{orange!40} 0.437 \\

\midrule
\rowcolor{gray!20}
 \multicolumn{17}{c}{\textit{Sampling-Based Distribution \& \textbf{w/o} Few-shot Steering}} \\
\midrule
\multirow{3}{*}{Llama-8B}
 &  No-CoT  & \cellcolor{Maroon1!5} 0.021 & \cellcolor{Blue1!7} 0.423 & \cellcolor{orange!47} 0.695 & \cellcolor{Maroon1!46} 0.357 & \cellcolor{Blue1!54} 0.158 & \cellcolor{orange!37} 0.398 & \cellcolor{Maroon1!0} 0.002 & \cellcolor{Blue1!30} 0.286 & \cellcolor{orange!43} 0.631 & \cellcolor{Maroon1!8} 0.097 & \cellcolor{Blue1!36} 0.273 & \cellcolor{orange!44} 0.564 & \cellcolor{Maroon1!8} 0.027 & \cellcolor{Blue1!29} 0.358 & \cellcolor{orange!55} 0.521 \\
 &  CoT     & \cellcolor{Maroon1!16} 0.063 & \cellcolor{Blue1!2} 0.440 & \cellcolor{orange!48} 0.699 & \cellcolor{Maroon1!28} 0.215 & \cellcolor{Blue1!41} 0.207 & \cellcolor{orange!31} 0.355 & \cellcolor{Maroon1!13} 0.061 & \cellcolor{Blue1!29} 0.289 & \cellcolor{orange!43} 0.631 & \cellcolor{Maroon1!15} 0.143 & \cellcolor{Blue1!25} 0.308 & \cellcolor{orange!44} 0.566 & \cellcolor{Maroon1!0} 0.004 & \cellcolor{Blue1!25} 0.374 & \cellcolor{orange!51} 0.496 \\
 &  R1      & \cellcolor{Maroon1!31} 0.121 & \cellcolor{Blue1!-0} 0.447 & \cellcolor{orange!48} 0.697 & \cellcolor{Maroon1!19} 0.149 & \cellcolor{Blue1!34} 0.233 & \cellcolor{orange!27} 0.330 & \cellcolor{Maroon1!37} 0.169 & \cellcolor{Blue1!49} 0.232 & \cellcolor{orange!54} 0.690 & \cellcolor{Maroon1!7} 0.089 & \cellcolor{Blue1!24} 0.312 & \cellcolor{orange!52} 0.586 & \cellcolor{Maroon1!32} 0.099 & \cellcolor{Blue1!45} 0.292 & \cellcolor{orange!50} 0.494 \\
\midrule
\multirow{3}{*}{Qwen-14B}
 &  No-CoT  & \cellcolor{Maroon1!23} 0.090 & \cellcolor{Blue1!24} 0.361 & \cellcolor{orange!42} 0.669 & \cellcolor{Maroon1!17} 0.135 & \cellcolor{Blue1!42} 0.203 & \cellcolor{orange!31} 0.354 & \cellcolor{Maroon1!17} 0.080 & \cellcolor{Blue1!36} 0.271 & \cellcolor{orange!42} 0.629 & \cellcolor{Maroon1!1} 0.047 & \cellcolor{Blue1!17} 0.332 & \cellcolor{orange!45} 0.567 & \cellcolor{Maroon1!9} 0.031 & \cellcolor{Blue1!23} 0.382 & \cellcolor{orange!38} 0.426 \\
 &  CoT     & \cellcolor{Maroon1!18} 0.070 & \cellcolor{Blue1!36} 0.318 & \cellcolor{orange!46} 0.688 & \cellcolor{Maroon1!26} 0.202 & \cellcolor{Blue1!40} 0.210 & \cellcolor{orange!30} 0.350 & \cellcolor{Maroon1!21} 0.098 & \cellcolor{Blue1!37} 0.267 & \cellcolor{orange!46} 0.649 & \cellcolor{Maroon1!6} 0.083 & \cellcolor{Blue1!20} 0.324 & \cellcolor{orange!55} 0.593 & \cellcolor{Maroon1!13} 0.043 & \cellcolor{Blue1!28} 0.361 & \cellcolor{orange!50} 0.495 \\
 &  R1      & \cellcolor{Maroon1!32} 0.124 & \cellcolor{Blue1!46} 0.282 & \cellcolor{orange!49} 0.705 & \cellcolor{Maroon1!37} 0.287 & \cellcolor{Blue1!52} 0.165 & \cellcolor{orange!38} 0.406 & \cellcolor{Maroon1!32} 0.145 & \cellcolor{Blue1!43} 0.250 & \cellcolor{orange!53} 0.686 & \cellcolor{Maroon1!29} 0.234 & \cellcolor{Blue1!34} 0.281 & \cellcolor{orange!55} 0.595 & \cellcolor{Maroon1!15} 0.050 & \cellcolor{Blue1!41} 0.306 & \cellcolor{orange!46} 0.469 \\
\midrule
\multirow{3}{*}{Qwen-32B}
 &  No-CoT  & \cellcolor{Maroon1!23} 0.091 & \cellcolor{Blue1!27} 0.348 & \cellcolor{orange!49} 0.702 & \cellcolor{Maroon1!18} 0.142 & \cellcolor{Blue1!46} 0.187 & \cellcolor{orange!34} 0.376 & \cellcolor{Maroon1!20} 0.092 & \cellcolor{Blue1!38} 0.264 & \cellcolor{orange!41} 0.623 & \cellcolor{Maroon1!12} 0.124 & \cellcolor{Blue1!28} 0.297 & \cellcolor{orange!54} 0.590 & \cellcolor{Maroon1!13} 0.042 & \cellcolor{Blue1!27} 0.366 & \cellcolor{orange!33} 0.402 \\
 &  CoT     & \cellcolor{Maroon1!30} 0.118 & \cellcolor{Blue1!44} 0.287 & \cellcolor{orange!49} 0.702 & \cellcolor{Maroon1!36} 0.280 & \cellcolor{Blue1!52} 0.165 & \cellcolor{orange!42} 0.430 & \cellcolor{Maroon1!34} 0.157 & \cellcolor{Blue1!43} 0.251 & \cellcolor{orange!42} 0.627 & \cellcolor{Maroon1!25} 0.208 & \cellcolor{Blue1!31} 0.290 & \cellcolor{orange!53} 0.589 & \cellcolor{Maroon1!7} 0.025 & \cellcolor{Blue1!31} 0.349 & \cellcolor{orange!44} 0.458 \\
 &  R1      & \cellcolor{Maroon1!19} 0.073 & \cellcolor{Blue1!42} 0.294 & \cellcolor{orange!60} 0.759 & \cellcolor{Maroon1!32} 0.244 & \cellcolor{Blue1!51} 0.169 & \cellcolor{orange!39} 0.414 & \cellcolor{Maroon1!40} 0.184 & \cellcolor{Blue1!49} 0.233 & \cellcolor{orange!53} 0.685 & \cellcolor{Maroon1!22} 0.192 & \cellcolor{Blue1!32} 0.285 & \cellcolor{orange!60} 0.607 & \cellcolor{Maroon1!22} 0.071 & \cellcolor{Blue1!43} 0.301 & \cellcolor{orange!41} 0.442 \\
\midrule
\multirow{3}{*}{Llama-70B}
 &  No-CoT  & \cellcolor{Maroon1!6} 0.024 & \cellcolor{Blue1!10} 0.412 & \cellcolor{orange!43} 0.673 & \cellcolor{Maroon1!9} 0.074 & \cellcolor{Blue1!27} 0.263 & \cellcolor{orange!22} 0.298 & \cellcolor{Maroon1!1} 0.006 & \cellcolor{Blue1!29} 0.291 & \cellcolor{orange!45} 0.644 & \cellcolor{Maroon1!0} 0.043 & \cellcolor{Blue1!6} 0.367 & \cellcolor{orange!44} 0.565 & \cellcolor{Maroon1!3} 0.014 & \cellcolor{Blue1!20} 0.393 & \cellcolor{orange!54} 0.513 \\
 &  CoT     & \cellcolor{Maroon1!32} 0.124 & \cellcolor{Blue1!25} 0.357 & \cellcolor{orange!47} 0.693 & \cellcolor{Maroon1!19} 0.146 & \cellcolor{Blue1!39} 0.216 & \cellcolor{orange!28} 0.337 & \cellcolor{Maroon1!10} 0.046 & \cellcolor{Blue1!29} 0.289 & \cellcolor{orange!46} 0.649 & \cellcolor{Maroon1!1} 0.053 & \cellcolor{Blue1!8} 0.361 & \cellcolor{orange!42} 0.560 & \cellcolor{Maroon1!9} 0.030 & \cellcolor{Blue1!30} 0.355 & \cellcolor{orange!54} 0.516 \\
 &  R1      & \cellcolor{Maroon1!23} 0.091 & \cellcolor{Blue1!47} 0.278 & \cellcolor{orange!58} 0.751 & \cellcolor{Maroon1!22} 0.175 & \cellcolor{Blue1!41} 0.208 & \cellcolor{orange!29} 0.344 & \cellcolor{Maroon1!34} 0.158 & \cellcolor{Blue1!46} 0.240 & \cellcolor{orange!56} 0.699 & \cellcolor{Maroon1!10} 0.112 & \cellcolor{Blue1!23} 0.313 & \cellcolor{orange!54} 0.591 & \cellcolor{Maroon1!20} 0.063 & \cellcolor{Blue1!39} 0.315 & \cellcolor{orange!48} 0.484 \\

\midrule
\rowcolor{gray!20}
 \multicolumn{17}{c}{\textit{Sampling-Based Distribution + Few-shot Steering}} \\
\midrule
\multirow{3}{*}{Llama-8B}
 &  No-CoT  & \cellcolor{Maroon1!0} 0.003 & \cellcolor{Blue1!9} 0.414 & \cellcolor{orange!48} 0.698 & \cellcolor{Maroon1!0} 0.004 & \cellcolor{Blue1!14} 0.313 & \cellcolor{orange!16} 0.257 & \cellcolor{Maroon1!14} 0.064 & \cellcolor{Blue1!-0} 0.373 & \cellcolor{orange!30} 0.563 & \cellcolor{Maroon1!8} 0.097 & \cellcolor{Blue1!-0} 0.386 & \cellcolor{orange!28} 0.522 & \cellcolor{Maroon1!21} 0.067 & \cellcolor{Blue1!-0} 0.476 & \cellcolor{orange!52} 0.504 \\
 &  CoT     & \cellcolor{Maroon1!1} 0.006 & \cellcolor{Blue1!2} 0.440 & \cellcolor{orange!48} 0.697 & \cellcolor{Maroon1!19} 0.150 & \cellcolor{Blue1!33} 0.237 & \cellcolor{orange!27} 0.332 & \cellcolor{Maroon1!15} 0.070 & \cellcolor{Blue1!34} 0.275 & \cellcolor{orange!46} 0.646 & \cellcolor{Maroon1!8} 0.098 & \cellcolor{Blue1!19} 0.326 & \cellcolor{orange!44} 0.565 & \cellcolor{Maroon1!28} 0.088 & \cellcolor{Blue1!43} 0.299 & \cellcolor{orange!17} 0.313 \\
 &  R1      & \cellcolor{Maroon1!5} 0.022 & \cellcolor{Blue1!1} 0.445 & \cellcolor{orange!48} 0.699 & \cellcolor{Maroon1!14} 0.114 & \cellcolor{Blue1!34} 0.236 & \cellcolor{orange!28} 0.339 & \cellcolor{Maroon1!40} 0.182 & \cellcolor{Blue1!51} 0.227 & \cellcolor{orange!54} 0.689 & \cellcolor{Maroon1!21} 0.181 & \cellcolor{Blue1!35} 0.275 & \cellcolor{orange!60} 0.607 & \cellcolor{Maroon1!19} 0.060 & \cellcolor{Blue1!45} 0.290 & \cellcolor{orange!48} 0.483 \\
\midrule
\multirow{3}{*}{Qwen-14B}
 &  No-CoT  & \cellcolor{Maroon1!21} 0.084 & \cellcolor{Blue1!25} 0.357 & \cellcolor{orange!45} 0.685 & \cellcolor{Maroon1!19} 0.151 & \cellcolor{Blue1!41} 0.208 & \cellcolor{orange!30} 0.348 & \cellcolor{Maroon1!19} 0.087 & \cellcolor{Blue1!26} 0.298 & \cellcolor{orange!43} 0.634 & \cellcolor{Maroon1!7} 0.087 & \cellcolor{Blue1!21} 0.320 & \cellcolor{orange!46} 0.570 & \cellcolor{Maroon1!27} 0.084 & \cellcolor{Blue1!14} 0.417 & \cellcolor{orange!52} 0.504 \\
 &  CoT     & \cellcolor{Maroon1!16} 0.062 & \cellcolor{Blue1!36} 0.316 & \cellcolor{orange!48} 0.697 & \cellcolor{Maroon1!34} 0.266 & \cellcolor{Blue1!50} 0.175 & \cellcolor{orange!37} 0.394 & \cellcolor{Maroon1!26} 0.121 & \cellcolor{Blue1!32} 0.282 & \cellcolor{orange!46} 0.646 & \cellcolor{Maroon1!14} 0.139 & \cellcolor{Blue1!20} 0.324 & \cellcolor{orange!49} 0.579 & \cellcolor{Maroon1!11} 0.037 & \cellcolor{Blue1!35} 0.333 & \cellcolor{orange!0} 0.222 \\
 &  R1      & \cellcolor{Maroon1!31} 0.121 & \cellcolor{Blue1!43} 0.290 & \cellcolor{orange!47} 0.692 & \cellcolor{Maroon1!42} 0.322 & \cellcolor{Blue1!54} 0.158 & \cellcolor{orange!36} 0.389 & \cellcolor{Maroon1!30} 0.137 & \cellcolor{Blue1!40} 0.257 & \cellcolor{orange!51} 0.673 & \cellcolor{Maroon1!25} 0.209 & \cellcolor{Blue1!34} 0.281 & \cellcolor{orange!58} 0.601 & \cellcolor{Maroon1!21} 0.068 & \cellcolor{Blue1!40} 0.310 & \cellcolor{orange!49} 0.488 \\
\midrule
\multirow{3}{*}{Qwen-32B}
 &  No-CoT  & \cellcolor{Maroon1!26} 0.101 & \cellcolor{Blue1!18} 0.381 & \cellcolor{orange!46} 0.687 & \cellcolor{Maroon1!18} 0.142 & \cellcolor{Blue1!47} 0.183 & \cellcolor{orange!34} 0.375 & \cellcolor{Maroon1!24} 0.111 & \cellcolor{Blue1!38} 0.263 & \cellcolor{orange!46} 0.646 & \cellcolor{Maroon1!10} 0.111 & \cellcolor{Blue1!27} 0.301 & \cellcolor{orange!52} 0.585 & \cellcolor{Maroon1!10} 0.034 & \cellcolor{Blue1!25} 0.372 & \cellcolor{orange!50} 0.493 \\
 &  CoT     & \cellcolor{Maroon1!34} 0.130 & \cellcolor{Blue1!46} 0.281 & \cellcolor{orange!50} 0.709 & \cellcolor{Maroon1!35} 0.272 & \cellcolor{Blue1!52} 0.166 & \cellcolor{orange!40} 0.416 & \cellcolor{Maroon1!26} 0.120 & \cellcolor{Blue1!42} 0.253 & \cellcolor{orange!48} 0.661 & \cellcolor{Maroon1!10} 0.111 & \cellcolor{Blue1!21} 0.320 & \cellcolor{orange!44} 0.564 & \cellcolor{Maroon1!16} 0.051 & \cellcolor{Blue1!36} 0.330 & \cellcolor{orange!25} 0.358 \\
 &  R1      & \cellcolor{Maroon1!4} 0.019 & \cellcolor{Blue1!38} 0.308 & \cellcolor{orange!57} 0.743 & \cellcolor{Maroon1!32} 0.246 & \cellcolor{Blue1!52} 0.164 & \cellcolor{orange!40} 0.419 & \cellcolor{Maroon1!38} 0.174 & \cellcolor{Blue1!47} 0.237 & \cellcolor{orange!56} 0.701 & \cellcolor{Maroon1!18} 0.161 & \cellcolor{Blue1!31} 0.290 & \cellcolor{orange!59} 0.604 & \cellcolor{Maroon1!27} 0.084 & \cellcolor{Blue1!43} 0.299 & \cellcolor{orange!46} 0.473 \\
\midrule
\multirow{3}{*}{Llama-70B}
 &  No-CoT  & \cellcolor{Maroon1!6} 0.025 & \cellcolor{Blue1!4} 0.433 & \cellcolor{orange!49} 0.703 & \cellcolor{Maroon1!2} 0.018 & \cellcolor{Blue1!35} 0.231 & \cellcolor{orange!28} 0.335 & \cellcolor{Maroon1!19} 0.090 & \cellcolor{Blue1!25} 0.300 & \cellcolor{orange!46} 0.646 & \cellcolor{Maroon1!12} 0.120 & \cellcolor{Blue1!19} 0.326 & \cellcolor{orange!55} 0.593 & \cellcolor{Maroon1!6} 0.023 & \cellcolor{Blue1!9} 0.438 & \cellcolor{orange!52} 0.505 \\
 &  CoT     & \cellcolor{Maroon1!20} 0.077 & \cellcolor{Blue1!35} 0.322 & \cellcolor{orange!51} 0.715 & \cellcolor{Maroon1!20} 0.158 & \cellcolor{Blue1!45} 0.192 & \cellcolor{orange!36} 0.391 & \cellcolor{Maroon1!4} 0.022 & \cellcolor{Blue1!24} 0.303 & \cellcolor{orange!45} 0.644 & \cellcolor{Maroon1!8} 0.098 & \cellcolor{Blue1!20} 0.323 & \cellcolor{orange!54} 0.590 & \cellcolor{Maroon1!32} 0.100 & \cellcolor{Blue1!36} 0.329 & \cellcolor{orange!31} 0.389 \\
 &  R1      & \cellcolor{Maroon1!16} 0.063 & \cellcolor{Blue1!44} 0.288 & \cellcolor{orange!58} 0.749 & \cellcolor{Maroon1!30} 0.234 & \cellcolor{Blue1!47} 0.184 & \cellcolor{orange!36} 0.388 & \cellcolor{Maroon1!32} 0.148 & \cellcolor{Blue1!44} 0.247 & \cellcolor{orange!53} 0.687 & \cellcolor{Maroon1!23} 0.197 & \cellcolor{Blue1!28} 0.299 & \cellcolor{orange!54} 0.592 & \cellcolor{Maroon1!22} 0.069 & \cellcolor{Blue1!38} 0.320 & \cellcolor{orange!47} 0.475 \\
\bottomrule
\end{tabular}

%% file: figures/tab_appendix_dis_colored.tex
\begin{tabular}{llccccc}
\toprule
 & & HS2$\downarrow$ & GHC$\downarrow$ & Pos$\downarrow$ & Neg$\downarrow$ & Amb$\downarrow$ \\
\midrule
\rowcolor{gray!20}
 \multicolumn{7}{c}{\textit{Verbalized Distribution \& \textbf{w/o} Few-shot Steering}} \\
\midrule
\multirow{3}{*}{Llama-8B}
 &  No-CoT  & \cellcolor{Blue1!58} 0.182 & \cellcolor{Blue1!33} 0.317 & \cellcolor{Blue1!38} 0.284 & \cellcolor{Blue1!36} 0.296 & \cellcolor{Blue1!58} 0.165 \\
 &  CoT     & \cellcolor{Blue1!58} 0.178 & \cellcolor{Blue1!52} 0.222 & \cellcolor{Blue1!54} 0.205 & \cellcolor{Blue1!48} 0.229 & \cellcolor{Blue1!60} 0.156 \\
 &  R1      & \cellcolor{Blue1!53} 0.204 & \cellcolor{Blue1!40} 0.280 & \cellcolor{Blue1!42} 0.263 & \cellcolor{Blue1!36} 0.291 & \cellcolor{Blue1!46} 0.232 \\
\midrule
\multirow{3}{*}{Qwen-14B}
 &  No-CoT  & \cellcolor{Blue1!47} 0.236 & \cellcolor{Blue1!37} 0.293 & \cellcolor{Blue1!28} 0.328 & \cellcolor{Blue1!32} 0.318 & \cellcolor{Blue1!41} 0.258 \\
 &  CoT     & \cellcolor{Blue1!48} 0.230 & \cellcolor{Blue1!56} 0.200 & \cellcolor{Blue1!35} 0.295 & \cellcolor{Blue1!46} 0.239 & \cellcolor{Blue1!45} 0.235 \\
 &  R1      & \cellcolor{Blue1!51} 0.216 & \cellcolor{Blue1!49} 0.235 & \cellcolor{Blue1!38} 0.284 & \cellcolor{Blue1!42} 0.262 & \cellcolor{Blue1!37} 0.283 \\
\midrule
\multirow{3}{*}{Qwen-32B}
 &  No-CoT  & \cellcolor{Blue1!43} 0.253 & \cellcolor{Blue1!48} 0.240 & \cellcolor{Blue1!33} 0.303 & \cellcolor{Blue1!49} 0.222 & \cellcolor{Blue1!41} 0.261 \\
 &  CoT     & \cellcolor{Blue1!45} 0.242 & \cellcolor{Blue1!56} 0.199 & \cellcolor{Blue1!44} 0.252 & \cellcolor{Blue1!58} 0.173 & \cellcolor{Blue1!47} 0.226 \\
 &  R1      & \cellcolor{Blue1!48} 0.227 & \cellcolor{Blue1!48} 0.242 & \cellcolor{Blue1!38} 0.281 & \cellcolor{Blue1!42} 0.257 & \cellcolor{Blue1!36} 0.284 \\
\midrule
\multirow{3}{*}{Llama-70B}
 &  No-CoT  & \cellcolor{Blue1!35} 0.294 & \cellcolor{Blue1!44} 0.262 & \cellcolor{Blue1!33} 0.307 & \cellcolor{Blue1!39} 0.277 & \cellcolor{Blue1!47} 0.225 \\
 &  CoT     & \cellcolor{Blue1!60} 0.170 & \cellcolor{Blue1!60} 0.180 & \cellcolor{Blue1!53} 0.210 & \cellcolor{Blue1!51} 0.207 & \cellcolor{Blue1!58} 0.165 \\
 &  R1      & \cellcolor{Blue1!47} 0.235 & \cellcolor{Blue1!49} 0.236 & \cellcolor{Blue1!43} 0.257 & \cellcolor{Blue1!43} 0.255 & \cellcolor{Blue1!45} 0.235 \\
\midrule
\multirow{3}{*}{Deepseek}
 &  V3-no-CoT  & \cellcolor{Blue1!54} 0.199 & \cellcolor{Blue1!46} 0.248 & \cellcolor{Blue1!45} 0.249 & \cellcolor{Blue1!38} 0.282 & \cellcolor{Blue1!50} 0.210 \\
 &  V3-CoT     & \cellcolor{Blue1!51} 0.217 & \cellcolor{Blue1!55} 0.207 & \cellcolor{Blue1!50} 0.223 & \cellcolor{Blue1!46} 0.237 & \cellcolor{Blue1!55} 0.184 \\
 &  R1         & \cellcolor{Blue1!48} 0.227 & \cellcolor{Blue1!55} 0.206 & \cellcolor{Blue1!52} 0.217 & \cellcolor{Blue1!46} 0.239 & \cellcolor{Blue1!53} 0.195 \\
\midrule
\rowcolor{gray!20}
 \multicolumn{7}{c}{\textit{Verbalized Distribution + Few-shot Steering}} \\
\midrule
\multirow{3}{*}{Llama-8B}
 &  No-CoT  & \cellcolor{Blue1!49} 0.225 & \cellcolor{Blue1!41} 0.274 & \cellcolor{Blue1!60} 0.178 & \cellcolor{Blue1!55} 0.188 & \cellcolor{Blue1!51} 0.204 \\
 &  CoT     & \cellcolor{Blue1!43} 0.254 & \cellcolor{Blue1!51} 0.226 & \cellcolor{Blue1!51} 0.222 & \cellcolor{Blue1!47} 0.232 & \cellcolor{Blue1!59} 0.159 \\
 &  R1      & \cellcolor{Blue1!43} 0.255 & \cellcolor{Blue1!49} 0.234 & \cellcolor{Blue1!42} 0.263 & \cellcolor{Blue1!39} 0.276 & \cellcolor{Blue1!38} 0.276 \\
\midrule
\multirow{3}{*}{Qwen-14B}
 &  No-CoT  & \cellcolor{Blue1!22} 0.357 & \cellcolor{Blue1!58} 0.188 & \cellcolor{Blue1!49} 0.231 & \cellcolor{Blue1!50} 0.213 & \cellcolor{Blue1!44} 0.245 \\
 &  CoT     & \cellcolor{Blue1!36} 0.289 & \cellcolor{Blue1!57} 0.193 & \cellcolor{Blue1!40} 0.271 & \cellcolor{Blue1!46} 0.240 & \cellcolor{Blue1!37} 0.278 \\
 &  R1      & \cellcolor{Blue1!44} 0.251 & \cellcolor{Blue1!49} 0.236 & \cellcolor{Blue1!40} 0.270 & \cellcolor{Blue1!43} 0.255 & \cellcolor{Blue1!36} 0.286 \\
\midrule
\multirow{3}{*}{Qwen-32B}
 &  No-CoT  & \cellcolor{Blue1!30} 0.317 & \cellcolor{Blue1!50} 0.232 & \cellcolor{Blue1!47} 0.240 & \cellcolor{Blue1!60} 0.159 & \cellcolor{Blue1!41} 0.259 \\
 &  CoT     & \cellcolor{Blue1!32} 0.307 & \cellcolor{Blue1!55} 0.203 & \cellcolor{Blue1!47} 0.239 & \cellcolor{Blue1!54} 0.193 & \cellcolor{Blue1!32} 0.305 \\
 &  R1      & \cellcolor{Blue1!25} 0.341 & \cellcolor{Blue1!48} 0.239 & \cellcolor{Blue1!39} 0.278 & \cellcolor{Blue1!40} 0.270 & \cellcolor{Blue1!22} 0.360 \\
\midrule
\multirow{3}{*}{Llama-70B}
 &  No-CoT  & \cellcolor{Blue1!33} 0.306 & \cellcolor{Blue1!43} 0.266 & \cellcolor{Blue1!35} 0.296 & \cellcolor{Blue1!40} 0.269 & \cellcolor{Blue1!43} 0.246 \\
 &  CoT     & \cellcolor{Blue1!43} 0.256 & \cellcolor{Blue1!54} 0.209 & \cellcolor{Blue1!55} 0.202 & \cellcolor{Blue1!53} 0.196 & \cellcolor{Blue1!57} 0.173 \\
 &  R1      & \cellcolor{Blue1!39} 0.273 & \cellcolor{Blue1!46} 0.249 & \cellcolor{Blue1!40} 0.272 & \cellcolor{Blue1!40} 0.271 & \cellcolor{Blue1!40} 0.262 \\
\midrule
\multirow{3}{*}{Deepseek}
 &  V3-no-CoT  & \cellcolor{Blue1!51} 0.216 & \cellcolor{Blue1!52} 0.218 & \cellcolor{Blue1!51} 0.219 & \cellcolor{Blue1!34} 0.305 & \cellcolor{Blue1!50} 0.210 \\
 &  V3-CoT     & \cellcolor{Blue1!36} 0.288 & \cellcolor{Blue1!51} 0.226 & \cellcolor{Blue1!45} 0.251 & \cellcolor{Blue1!33} 0.309 & \cellcolor{Blue1!44} 0.241 \\
 &  R1         & \cellcolor{Blue1!32} 0.308 & \cellcolor{Blue1!55} 0.204 & \cellcolor{Blue1!52} 0.218 & \cellcolor{Blue1!48} 0.228 & \cellcolor{Blue1!46} 0.231 \\

\midrule
\rowcolor{gray!20}
 \multicolumn{7}{c}{\textit{Sampling-Based Distribution \& \textbf{w/o} Few-shot Steering}} \\
\midrule
\multirow{3}{*}{Llama-8B}
 &  No-CoT  & \cellcolor{Blue1!12} 0.408 & \cellcolor{Blue1!30} 0.333 & \cellcolor{Blue1!40} 0.274 & \cellcolor{Blue1!28} 0.339 & \cellcolor{Blue1!44} 0.240 \\
 &  CoT     & \cellcolor{Blue1!5} 0.440 & \cellcolor{Blue1!23} 0.365 & \cellcolor{Blue1!25} 0.341 & \cellcolor{Blue1!20} 0.381 & \cellcolor{Blue1!31} 0.315 \\
 &  R1      & \cellcolor{Blue1!1} 0.461 & \cellcolor{Blue1!19} 0.386 & \cellcolor{Blue1!27} 0.334 & \cellcolor{Blue1!16} 0.405 & \cellcolor{Blue1!38} 0.274 \\
\midrule
\multirow{3}{*}{Qwen-14B}
 &  No-CoT  & \cellcolor{Blue1!7} 0.433 & \cellcolor{Blue1!1} 0.476 & \cellcolor{Blue1!2} 0.451 & \cellcolor{Blue1!1} 0.492 & \cellcolor{Blue1!6} 0.447 \\
 &  CoT     & \cellcolor{Blue1!34} 0.298 & \cellcolor{Blue1!16} 0.402 & \cellcolor{Blue1!14} 0.397 & \cellcolor{Blue1!10} 0.437 & \cellcolor{Blue1!23} 0.354 \\
 &  R1      & \cellcolor{Blue1!35} 0.293 & \cellcolor{Blue1!18} 0.389 & \cellcolor{Blue1!17} 0.381 & \cellcolor{Blue1!14} 0.415 & \cellcolor{Blue1!26} 0.338 \\
\midrule
\multirow{3}{*}{Qwen-32B}
 &  No-CoT  & \cellcolor{Blue1!8} 0.429 & \cellcolor{Blue1!2} 0.469 & \cellcolor{Blue1!3} 0.449 & \cellcolor{Blue1!4} 0.474 & \cellcolor{Blue1!7} 0.442 \\
 &  CoT     & \cellcolor{Blue1!28} 0.327 & \cellcolor{Blue1!13} 0.417 & \cellcolor{Blue1!13} 0.400 & \cellcolor{Blue1!12} 0.427 & \cellcolor{Blue1!20} 0.372 \\
 &  R1      & \cellcolor{Blue1!24} 0.349 & \cellcolor{Blue1!17} 0.398 & \cellcolor{Blue1!18} 0.375 & \cellcolor{Blue1!13} 0.422 & \cellcolor{Blue1!27} 0.336 \\
\midrule
\multirow{3}{*}{Llama-70B}
 &  No-CoT  & \cellcolor{Blue1!-0} 0.467 & \cellcolor{Blue1!1} 0.478 & \cellcolor{Blue1!3} 0.446 & \cellcolor{Blue1!-0} 0.495 & \cellcolor{Blue1!6} 0.451 \\
 &  CoT     & \cellcolor{Blue1!26} 0.338 & \cellcolor{Blue1!10} 0.430 & \cellcolor{Blue1!13} 0.400 & \cellcolor{Blue1!5} 0.469 & \cellcolor{Blue1!19} 0.379 \\
 &  R1      & \cellcolor{Blue1!31} 0.316 & \cellcolor{Blue1!9} 0.434 & \cellcolor{Blue1!17} 0.379 & \cellcolor{Blue1!9} 0.443 & \cellcolor{Blue1!24} 0.353 \\

\midrule
\rowcolor{gray!20}
 \multicolumn{7}{c}{\textit{Sampling-Based Distribution + Few-shot Steering}} \\
\midrule
\multirow{3}{*}{Llama-8B}
 &  No-CoT  & \cellcolor{Blue1!18} 0.380 & \cellcolor{Blue1!18} 0.393 & \cellcolor{Blue1!23} 0.353 & \cellcolor{Blue1!19} 0.389 & \cellcolor{Blue1!18} 0.384 \\
 &  CoT     & \cellcolor{Blue1!6} 0.435 & \cellcolor{Blue1!20} 0.383 & \cellcolor{Blue1!25} 0.342 & \cellcolor{Blue1!18} 0.392 & \cellcolor{Blue1!41} 0.259 \\
 &  R1      & \cellcolor{Blue1!4} 0.448 & \cellcolor{Blue1!18} 0.391 & \cellcolor{Blue1!24} 0.349 & \cellcolor{Blue1!20} 0.381 & \cellcolor{Blue1!36} 0.286 \\
\midrule
\multirow{3}{*}{Qwen-14B}
 &  No-CoT  & \cellcolor{Blue1!11} 0.415 & \cellcolor{Blue1!5} 0.456 & \cellcolor{Blue1!3} 0.447 & \cellcolor{Blue1!2} 0.483 & \cellcolor{Blue1!5} 0.453 \\
 &  CoT     & \cellcolor{Blue1!34} 0.297 & \cellcolor{Blue1!16} 0.403 & \cellcolor{Blue1!12} 0.403 & \cellcolor{Blue1!11} 0.436 & \cellcolor{Blue1!15} 0.398 \\
 &  R1      & \cellcolor{Blue1!29} 0.321 & \cellcolor{Blue1!20} 0.381 & \cellcolor{Blue1!16} 0.384 & \cellcolor{Blue1!14} 0.415 & \cellcolor{Blue1!28} 0.327 \\
\midrule
\multirow{3}{*}{Qwen-32B}
 &  No-CoT  & \cellcolor{Blue1!7} 0.430 & \cellcolor{Blue1!3} 0.465 & \cellcolor{Blue1!4} 0.443 & \cellcolor{Blue1!5} 0.469 & \cellcolor{Blue1!6} 0.451 \\
 &  CoT     & \cellcolor{Blue1!28} 0.330 & \cellcolor{Blue1!12} 0.419 & \cellcolor{Blue1!15} 0.389 & \cellcolor{Blue1!13} 0.420 & \cellcolor{Blue1!19} 0.379 \\
 &  R1      & \cellcolor{Blue1!22} 0.356 & \cellcolor{Blue1!16} 0.400 & \cellcolor{Blue1!19} 0.370 & \cellcolor{Blue1!13} 0.421 & \cellcolor{Blue1!28} 0.332 \\
\midrule
\multirow{3}{*}{Llama-70B}
 &  No-CoT  & \cellcolor{Blue1!2} 0.457 & \cellcolor{Blue1!-0} 0.481 & \cellcolor{Blue1!-0} 0.461 & \cellcolor{Blue1!2} 0.482 & \cellcolor{Blue1!-0} 0.481 \\
 &  CoT     & \cellcolor{Blue1!27} 0.333 & \cellcolor{Blue1!9} 0.434 & \cellcolor{Blue1!7} 0.427 & \cellcolor{Blue1!8} 0.449 & \cellcolor{Blue1!18} 0.385 \\
 &  R1      & \cellcolor{Blue1!29} 0.323 & \cellcolor{Blue1!11} 0.425 & \cellcolor{Blue1!16} 0.385 & \cellcolor{Blue1!13} 0.422 & \cellcolor{Blue1!22} 0.363 \\

\bottomrule
\end{tabular}